\pdfoutput=1

\documentclass[11pt]{article}

\usepackage[preprint]{acl}

\usepackage{times}
\usepackage{latexsym}
\usepackage{soul}

\usepackage[T1]{fontenc}

\usepackage[utf8]{inputenc}

\usepackage{microtype}

\usepackage{inconsolata}

\usepackage{amsmath}
\usepackage{multirow}
\usepackage{tabularx}
\usepackage{graphicx}
\usepackage{amssymb}
\usepackage{hyperref}
\usepackage{pifont}
\usepackage{algorithm}
\usepackage{algorithmic}
\usepackage[switch]{lineno}
\usepackage{booktabs}
\usepackage{bbding}
\usepackage{subfigure}
\usepackage{caption}
\usepackage{subcaption}
\usepackage{soul}
\usepackage{color, xcolor}
\usepackage{framed}
\usepackage{float}
\usepackage{makecell}

\newcommand{\red}[1]{\textcolor{red}{#1}}

\title{\textsc{ProTrix}: Building Models for Planning and Reasoning\\ over Tables with Sentence Context}

\author{~Zirui Wu,  ~Yansong Feng\textsuperscript{\Envelope}~~\\
 Wangxuan Institute of Computer Technology, Peking University, China \\
\texttt{\{ziruiwu,fengyansong\}@pku.edu.cn} \\
 }

\begin{document}
\maketitle
\begin{abstract}

Tables play a crucial role in conveying information of various domains. We propose a novel \textit{Plan-then-Reason} framework to answer different types of user queries over tables with sentence context. Our framework first plans the reasoning paths over both tabular and sentence context, then assigns each step to program-based or textual reasoning to reach the final answer. This framework enhances the table reasoning abilities for both in-context learning and finetuning methods. GPT-3.5-Turbo following our \textit{Plan-then-Reason} framework surpasses other prompting baselines that do not employ self-consistency while using fewer API calls and in-context demonstrations. We also construct an instruction tuning set \texttt{TrixInstruct} to evaluate the effectiveness of finetuning with this framework. We present a \textsc{ProTrix} model family by finetuning models on \texttt{TrixInstruct}. Our experiments show that the \textsc{ProTrix} family can generalize to diverse unseen tabular tasks with only 6k training instances. We further demonstrate that \textsc{ProTrix} can generate accurate and faithful explanations to answer complex free-form questions. Our work underscores the importance of the planning and reasoning abilities towards a model over tabular tasks with generalizability and interpretability. We open-source our dataset and models at \url{https://github.com/WilliamZR/ProTrix}.

\end{abstract}

\section{Introduction}
\label{sec:intro}
Tables, serve as a fundamental tool for organizing and presenting information across various domains. Whether in business reports, or scientific publications, tables are commonly employed to represent complex data effectively. Despite their widespread utility, the process of human beings answering questions involving tables appears to be time-consuming, given the often substantial amount of content involved. Recognizing this challenge, there arises a need to leverage Large Language Models (LLMs) to understand and respond to user queries automatically.
\begin{figure}[!t]
    \centering
    \includegraphics[width=1.0\linewidth]{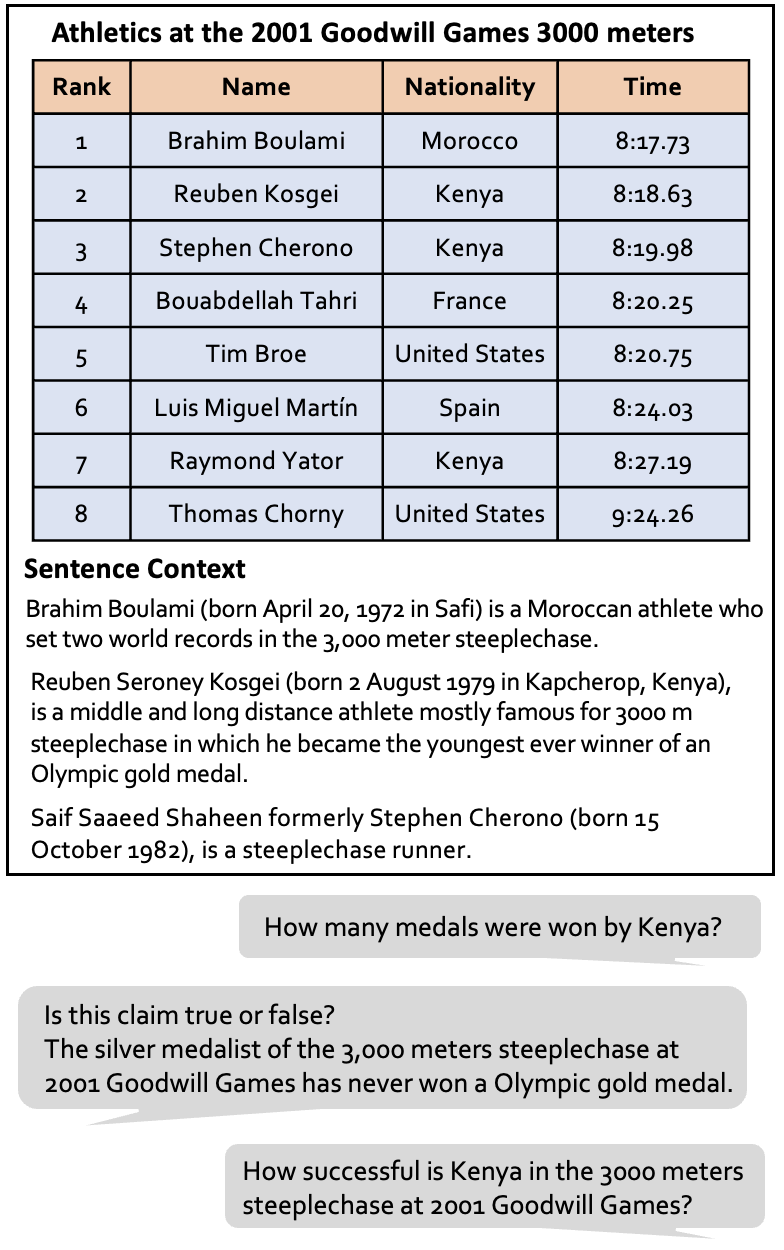}
    \caption{Demonstrations of user queries to a table in Wikipedia. Some of the sentences with hyperlinks to the table are presented as sentence context.}
    \label{fig:questions_demo}
\end{figure}

\begin{figure*}[!t]
    \centering
    \includegraphics[width=0.98\linewidth]{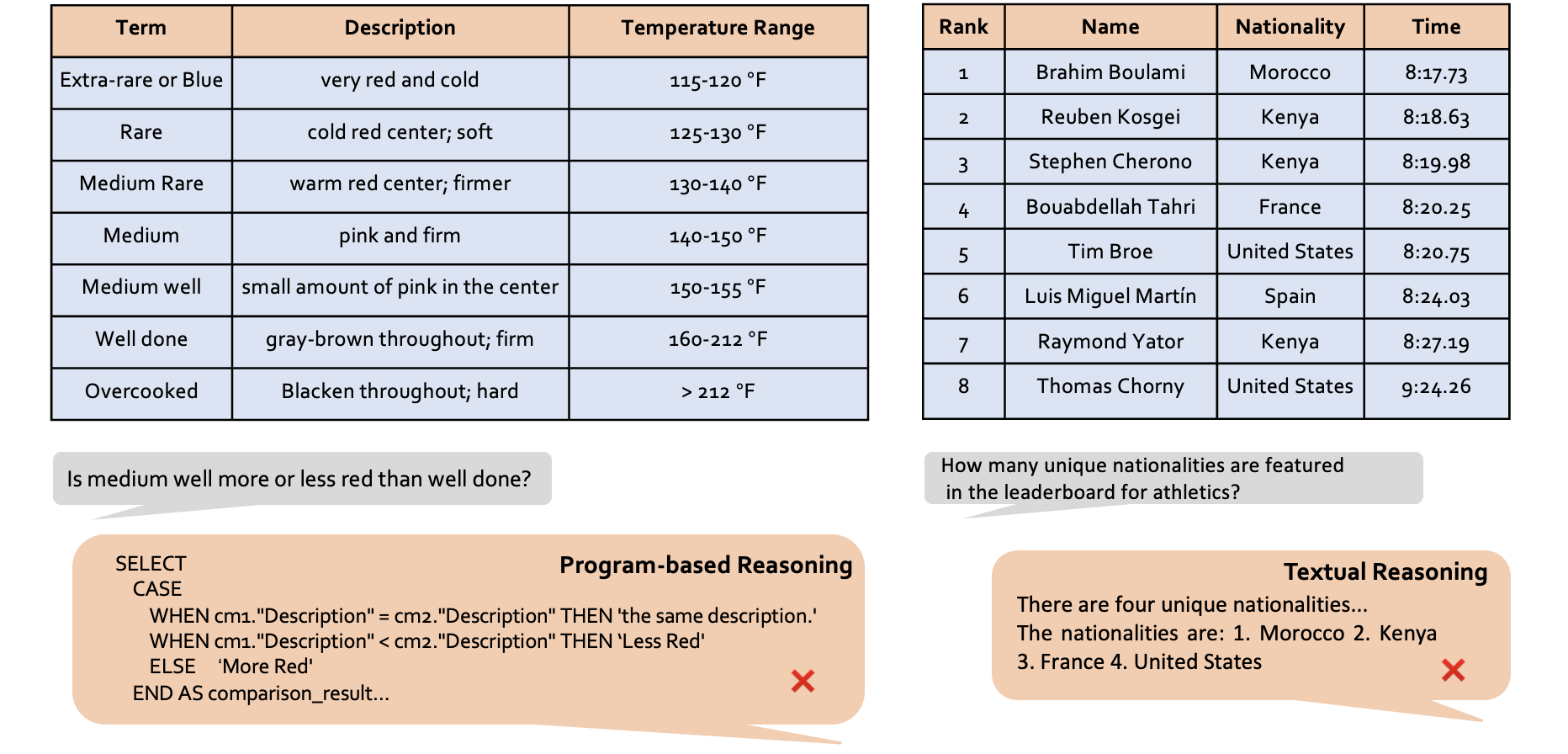}
    \caption{Demonstration of disadvantages of program-based and textual reasoning on tabular tasks. Program-based reasoning fails to answer the query since it tries to compare general concepts with a math operator. The textual reasoning fails on a program-solvable query that needs to count distinct countries in the table.}
    \label{fig:reasoning_demo}
\end{figure*}

Figure~\ref{fig:questions_demo} demonstrates three kinds of user queries for a table from Wikipedia. In the first example, the user query is \textit{how many medals were won by Kenya}. This question is annotated as a program-unsolvable question by SQL experts~\cite{shi2020squall} attributed to the absence of an explicit column for medals in the table. To resolve this, the model must fill the gap between the query and the table by recalling the commonsense knowledge that \textit{only the top three players can win medals}. 

The second query delves into a multi-hop scenario asking whether \textit{the silver medalist at the 2001 Goodwill Games has ever won an Olympic gold medal.} Addressing such queries raises two challenges (1) Decompose the query into sub-tasks. Such as the model plans to derive the silver medalists first and then verify their record of Olympic medals. (2) Combining structured and unstructured context. The model must extract the athletic name from the table and derive the information from the sentence context that Kosgei has won an Olympic gold medal since he is \textit{the youngest ever winner of an Olympic gold medal}.  

The last query also requires the model to recall commonsense knowledge to decide which contextual information can be used as evidence to judge if \textit{Kenya is successful at the 2001 Goodwill Games}. Subsequently, the model must generate explanations to arrive at certain conclusions. The first two queries mainly require the model to fill the information gap in the query with a short-form answer while the third query seeks for information on a more general concept. The queries underscore the importance of planning and reasoning abilities to connect the query with the actual information in the context and generate faithful and accurate explanations for conclusions.

There are generally two ways to enhance a model's reasoning ability. One is textual reasoning which prompts the model to answer questions step-by-step~\cite{wei2022chain}. The other one is program-based reasoning, prompting the model to write code to answer the questions~\cite{chen2022program}. However, both reasoning methods have their own disadvantages. The textual reasoning method such as Chain-of-Thought~\cite{wei2022chain} can be used to enhance the tabular reasoning ability but often lacks precision in tabular operations such as sorting, counting and filtering as shown in the right example in Figure~\ref{fig:reasoning_demo}, and may not generalize well to large tables~\cite{chen2023large}. The program-based reasoning method can reason with high precision with SQL or Python code~\cite{chen2022program}. The left one in Figure~\ref{fig:reasoning_demo} queries the color comparison between steaks with different cooking methods which is program-unsolvable by SQL and Python. 
Therefore, it requires the model to adaptively plan the usage of the program-based and textual reasoning to make the best of both methods.

Previous works incorporating LLMs' dynamic planning abilities for table reasoning~\cite{yao2023react, zhang2023reactable,wang2024chain} primarily focus on selecting local actions during the reasoning process, rather than globally planning the entire reasoning strategy beforehand. In this paper, we propose a novel \textit{Plan-then-Reason} framework, which decomposes the question and adapts the reasoning chain based on the table and sentence context in advance, preventing the model from becoming trapped in localized reasoning steps~\cite{zhang2023crt}. Instead of fixed reasoning patterns~\cite{wei2022chain,chen2022program,wang2023plan,nahid2024tabsqlify}, our approach offers greater flexibility by adaptively assigning reasoning sub-tasks to either SQL-based or text-based methods, depending on the nature of the decomposed sub-questions. Experiments demonstrate that our \textit{Plan-then-Reason} framework generates more accurate answers with fewer API calls, surpassing existing prompting methods while requiring fewer in-context examples.

Meanwhile, few finetuning methods are designed to enhance both planning and reasoning abilities of open-source models while they are crucial for building tabular models with generalizability and interpretability.
Various pre-trained models are proposed for tabular tasks~\cite{yin2020tabert,wang2021tuta,iida2021tabbie,deng2022turl,yang2022tableformer,jiang2022omnitab,liu2021tapex}. But they are limited to specific query types and could not generalize well to unseen tasks. 
Models finetuned with respect to general tabular querying tasks~\cite{xie2022unifiedskg,liu2023zero,zhang2023tablellama, zhuang2024structlm} are expected to generate answers directly, which inevitably lacks interpretability.
The reasoning method for TableLLM~\cite{zhang2024tablellm} is primarily determined by the size of tables, rather than being planned based on the query and context.

Recent base models are pre-trained with a large amount of corpora thus obtaining intrinsic commonsense knowledge~\cite{touvron2023llama,roziere2023code}. These models suit as the backbones for our framework that can fill the gap between queries and tables, understand general concepts, and plan the reasoning paths over table and sentence context.
We construct an instruction tuning dataset \texttt{TrixInstruct} based on benchmarks with queries that are program-unsolvable or need combining information from table and sentence context. We finetune Llama-2-7B~\cite{touvron2023llama} and CodeLlama-7B~\cite{roziere2023code} with \texttt{TrixInstruct}. The resulting \textsc{ProTrix}\footnote{Protrix originally means a chemical reactor for small-scale production with compatibility and process control.} model family is designed to \underline{\textbf{P}}lan and \underline{\textbf{R}}eason \underline{\textbf{O}}n \underline{\textbf{T}}abula\underline{\textbf{R}} tasks with integration of code execut\underline{\textbf{I}}on and te\underline{\textbf{X}}tual reasoning.  Our experiments show that models trained with our \textit{Plan-then-Reason} framework can generalize to unseen tabular tasks in different domains with only a handful of training examples and give accurate and faithful explanations even for complex \textit{how} and \textit{why} questions.

In summary, our contributions are listed as:

\noindent $\bullet$ We propose a \textit{Plan-then-Reason} framework to answer user queries on tabular tasks with sentence context. The framework first plans the reasoning pathways by ingesting the query and the context, and assigns each step to textual and program-based reasoning to arrive at the final answer. We experiment with GPT-3.5-turbo to evaluate the effectiveness of this framework and find it surpasses existing methods that do not employ self-consistency.

\noindent $\bullet$ We construct \texttt{TrixInstruct}, an instruction-tuning set to build models with generalizability and interpretability over tables with sentence context. 
To obtain the required planning and reasoning abilities, we include queries that are program-unsolvable or need combining tables and sentences in our instruction-tuning dataset.

\noindent $\bullet$ We open-source our model \textsc{ProTrix}, capable of planning and reasoning on tabular tasks with sentence context. \textsc{ProTrix} can generalize to unseen tabular tasks with sentence context, and generate accurate and faithful explanations.

\section{Our Method}
\subsection{Problem Formulation}

This study centers on tabular tasks with sentence context.
Each instance is structured as $(Q, T, S, A)$, where $Q$ represents users' query. $T$ denotes a singular table, while $S$ denotes the sentence context. The sentence context is usually passages linked to the table or retrieved from a knowledge base. Finally $A$ stands for the answer, which could be short-form for answering questions like \textit{hom many...} or \textit{is this true or false...}. For \textit{how} and \textit{why} questions, the answers are generally one or more sentences, defined as free-form answers.

\begin{figure*}
    \centering
    \includegraphics[width=0.95\linewidth]{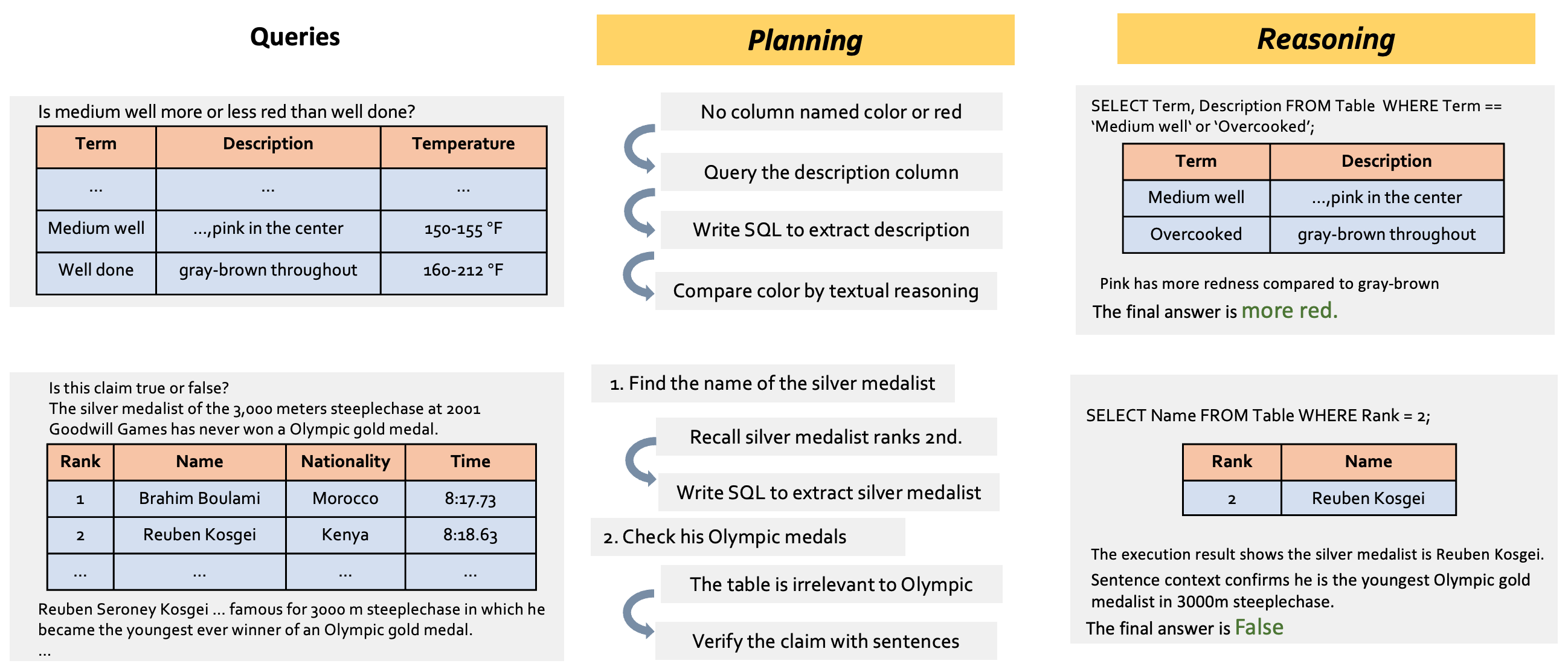}
    \caption{Illustration of our framework. The figure shows the process of our framework to answer a question. The framework first performs strategical planning, decomposing questions into 
    reasoning chains to be solved by either table or sentence context, then performs program-based or textual reasoning to reach the answers.}
    \label{fig:framework}
\end{figure*}

\subsection{Plan-then-Reason}
\label{sec:architecture}
 We propose a \textit{Plan-then-Reason} framework to answer queries over tables and texts. The framework first ingests the query and the context by recalling common knowledge and general concepts. Then it begins to design the model's reasoning pathway, planning the utilization of program-based and textual reasoning to arrive at conclusions.

\noindent\paragraph{Planning} The model first analyzes the query and fills the potential gap between the query and the context. Consider the first query in Figure~\ref{fig:framework}, there is no explicit column of \textit{color} in the table. The model recalls commonsense that \textit{pink}, \textit{gray-brown} and other colors in the description column can be used to answer the question. Similarly, in the second query, the model recalls that \textit{only top 3 athletes can win medals}. 

Then the model adaptively plans the reasoning path with both program-based and textual reasoning to address the limitations of each individual reasoning method. For the first query, the model plans to use SQL to extract relevant cells from the table and make comparisons of concepts through textual reasoning. For the second query, the model decomposes the task into a multi-hop reasoning chain. It uses SQL to extract the silver medalist from the table and uses sentence context to verify his Olympic record.

\noindent\paragraph{Reasoning} The reasoning phase initiates with program-based reasoning, writing SQL queries to extract relevant cells or perform operations such as counting and sorting. 
After running SQL on a code interpreter, the results are fed back into the model's ongoing reasoning process. Subsequently, during textual reasoning, the model selects relevant sentences from noisy context to complement the table context. \textit{Reuben Kosgei... youngest ever winner of an Olympic gold medal} suggests that he has won an Olympic gold medal in his career. Finally, the model summarizes insights from program-based and textual reasoning to reach the answer.

\subsection{In-Context Learning}

One intuitive way to evaluate the effectiveness of our \textit{Plan-then-Reason} work is through in-context learning. We prompt LLMs to follow the planning and reasoning pattern in Figure~\ref{fig:framework}.
We experiment with GPT-3.5-turbo on WikiTQ~\cite{pasupat-liang-2015-compositional}, FEVEROUS~\cite{aly2021feverous} and TabFact~\cite{chen2019tabfact} following the prompts shown in Table~\ref{tab:wikitab_distill},~\ref{tab:feverous_distill} and~\ref{tab:tabfact_distill} in Appendix. 

\subsection{Instruction Tuning}
\label{subsec:sft}
Based on the analysis in \S\ref{sec:intro}, we highlight the abilities our \textit{Plan-then-Reason} framework possesses towards tabular tasks with sentence context.
(1) Understanding user's query: use parametric knowledge of commonsense insights and general concepts to analyze the relationship between the query and the context; (2) Adaptive planning: decompose a query into sub-questions and plan to answer these sub-questions with different types of  
context or design multi-hop reasoning paths for the query, and (3) Blending program-based and textual reasoning: synthesize the strength of each method to maximize reasoning performance. 

To train our model with such abilities, we construct an instruction tuning set \texttt{TrixInstruct} based on two benchmarks i.e., WikiTQ~\cite{pasupat-liang-2015-compositional} and FEVEROUS \cite{aly2021feverous}. WikiTQ involves table question-answering based on a single table, requiring multi-step reasoning 
and performing diverse data operations such as comparison, aggregation, and arithmetic computation. WikiTQ also contains cases that can not be solved by SQL programming solely~\cite{shi2020squall} which require extra textual reasoning as the left example in Figure~\ref{fig:reasoning_demo}. 
Finetuning with such cases equips our models with the ability to plan between textual or program-based reasoning.

On the other hand, FEVEROUS presents an open-domain fact verification challenge spanning both sentences and tables. We select samples containing precisely one table in their gold evidence set. Each case is presented with the corresponding table along with 5 sentences as contextual information. To introduce variability to the sentence context, we ensure the inclusion of gold sentence evidence and augment the context with noisy sentences retrieved from Wikipedia by a dense retriever~\cite{hu2023unifee}. Training examples on claim verification from FEVEROUS impart the ability to decompose claims and reason on each sub-claim with a specific table or sentence context.

For each task, we sample 4,000 instances from the training datasets. 
We employ GPT-4~\cite{achiam2023gpt} to generate responses according to the \textit{Plan-then-Reason} framework following the prompts in Table~\ref{tab:wikitab_distill} and~\ref{tab:feverous_distill} in Appendix. We filter out instances that GPT-4 predicts answers inconsistent with the original annotations.
This results in a training set comprising 3,157 cases from FEVEROUS and 2,866 cases from WikiTQ. We train our models based on Llama-2-7B \cite{touvron2023llama} and CodeLlama-7B~\cite{roziere2023code} resulting in two models \textsc{ProTrix} and \textsc{ProTrix-Coder}. 
\sethlcolor{gray}

\begin{table*}[th]
\small
\setlength{\tabcolsep}{4pt}
	\centering
	\begin{tabular}{lccccccc}
		\toprule
& WikiTQ & WikiSQL & TabFact & \textsc{SciTAB} &FEVEROUS & HybridQA &TATQA\\
 \midrule
  \textit{GPT-4} &  &  &  &  & & &\\

  \quad End-to-End QA & 72.9 & 75.8 & 71.5 & 57.1& 71.0& 64.1 & 80.8\\
  \midrule
   \textit{GPT-3.5-turbo} &  &  &  &  & & &\\

 \quad End-to-End QA & 51.8& 55.0 & 68.8&45.3 & \underline{61.0}& 55.1 & 59.1\\
\quad ReAcTable  & 52.4 & - & 73.1 & - & - & - & - \\
\quad Dater$^{*}$ & 52.8 & - & 78.0 & - & - & - & - \\
\quad Binder$^{*}$  & 56.7 & - &79.2 &-& - & - &-\\
\quad StructGPT  & 57.0 & 64.6 & \textbf{87.3} & - & - & - & - \\
\quad Chain-of-Table & 59.9 & - & 80.2 & - & - & - & - \\
\quad TabSQLify & 64.7 & 76.7 & 79.5 &  - & - & - & - \\
\quad Mix SC$^{*}$ & \textbf{73.1}  & - & - &  - & - & - & - \\
\quad Plan-then-Reason (Ours) & \underline{65.2} & -& \underline{83.5}&- &\textbf{65.8}  &- & -\\

 \midrule
\textit{Finetuned SOTA} & ~~63.3$^{\dagger}$ & ~~89.2$^{\dagger}$ & ~~90.8$^{\dagger}$& - & ~~75.9$^{\dagger}$ & ~~61.0$^{\dagger}$&~~74.5$^{\dagger}$ \\
 \midrule
 \textit{Llama-2-7B} & & & & &  \\
   \quad End-to-End QA & 21.4& 17.4 &48.6 &27.2& 47.1 & 27.6 & 28.7\\
   \quad TableLlama & 31.6 & 41.7 & ~~\textbf{82.6}$^{\dagger}$&29.2 & 56.8 & 33.3&38.3 \\
    \quad \textsc{ProTrix}(Ours) & ~~\textbf{56.2}$^{\dagger}$ & \textbf{67.4} & 71.6 & \textbf{45.0}& ~~\textbf{75.6}$^{\dagger}$ & \textbf{42.9}&\textbf{50.1}\\
    \midrule
  \textit{CodeLlama-7B} & & & & &  \\
    \quad End-to-End QA & 13.1 & 17.3 & 49.5&37.1& 43.0 &28.5 &28.4\\
\quad TableLLM & ~~52.9$^{\dagger}$& ~~65.3$^{\dagger}$& 57.1&24.7 & 60.0 &\textbf{53.7} & ~~\textbf{70.3}$^{\dagger}$\\
    \quad \textsc{ProTrix-Coder} (Ours) & ~~\textbf{57.8}$^{\dagger}$&  \textbf{72.3} &\textbf{ 70.6} &\textbf{41.2}& ~~\textbf{71.4}$^{\dagger}$ & 45.1  & 52.2\\

    \bottomrule
	\end{tabular}
	\caption{Experimental results on short-form question answering and fact verification tasks.  $^{\dagger}$ The model is trained on this evaluation benchmark.  $^{*}$: with self-consistency.}
	\label{tab:results}
\end{table*}
\section{Experiments}
\subsection{Benchmarks for Evaluation}
We use existing tabular benchmarks with different input and output configurations to evaluate the performance of our model on queries with short-form or free-form answers. We further divide existing benchmarks on short-form answer tasks into short-form question answering and fact verification following the category in Figure~\ref{fig:questions_demo}.
\noindent\paragraph{Short-form Question Answering} WikiSQL and WikiTQ are question answering 
benchmarks on tables from Wikipedia without sentence context~\cite{zhong2017seq2sql,pasupat-liang-2015-compositional}. HybridQA~\cite{chen2020hybridqa} requires models to answer questions based on both tables and sentences. We use retrieved sentences, admittedly noisy, from \citet{chen2020hybridqa} as the sentence context. 
TATQA~\cite{zhu2021tat} focuses on tables with sentence context from financial reports.

\noindent\paragraph{Fact Verification} We use the same method in \S\ref{subsec:sft} to construct the evaluation dataset for FEVEROUS~\cite{aly2021feverous}. TabFact~\cite{chen2019tabfact} verifies claims based on tables from Wikipedia. \textsc{SciTAB}~\cite{lu-etal-2023-scitab} focuses on tables from scientific papers. This benchmark requires compositional reasoning and commonsense knowledge.

\noindent\paragraph{Free-form Question Answering} FetaQA contains \textit{what} questions with multiple answers and \textit{how} and \textit{why} questions that require models to generate explanations~\cite{nan2022fetaqa}. The original FetaQA dataset has highlighted relevant cells, we turn to a more challenging and realistic scenario where the highlighted cells are not provided as input and the model will answer the question directly based on the complete table context. Since our model is only finetuned on short-form answer tasks, FetaQA can be utilized to further evaluate the interpretability and generalizability of our models.

\subsection{Short-form Answer Tasks}

\noindent \paragraph{Baselines} We choose the following baselines: (1) Closed-source model: We use the end-to-end QA performance of GPT-3.5-turbo and GPT-4 as baselines and compare our results on GPT-3.5-turbo with Binder~\cite{cheng2022binding}, ReAcTable~\cite{zhang2023reactable}, StructGPT~\cite{jiang-etal-2023-structgpt}, Dater~\cite{ye2023large}, Chain-of-Table~\cite{wang2024chain} and TabSQLify~\cite{nahid2024tabsqlify}. (2) Finetuned SOTA: Please refer to Appendix~\ref{app:detail} for details of finetuned SOTA methods. (3) 7B parameter models: We first compare our models with the zero-shot performance of base models, Llama-2-7B~\cite{touvron2023llama} and CodeLlama 7B~\cite{roziere2023code}. Then we compare with TableLlama~\cite{zhang2023tablellama} and TableLLM~\cite{zhang2024tablellm}, which are the most similar baselines to our models and share the same base models\footnote{We do not use StructLM~\cite{zhuang2024structlm} as baseline since it is finetuned on most of our evaluation benchmarks while we want to compare with existing models in terms of generalizability.}. TableLlama is a generalist model for end-to-end QA. TableLLM originally chooses textual or program-based reasoning for different benchmarks. To test the generalizability for different queries, we choose the reasoning method of TableLLM by table size. We use textual reasoning for tables less than 500 tokens and program-based reasoning for larger tables. 
We provide a detailed comparison with other closed-source models (Codex, PaLM2) and open-source model (Llama-3-8B) in Table~\ref{tab:detailed_results} in Appendix.

 \noindent \paragraph{Evaluation Metrics}
For fact verification, we match the predicted veracity label in the concluding sentence, and use the evaluator from \citet{cheng2022binding} to evaluate the performance of \textit{Plan-then-Reason} with GPT-3.5-turbo on WikiTQ. For 7B parameter models, we find that the heuristic matching evaluation for question-answering tasks would lead to overestimation or underestimation since our model is not finetuned to follow the grammar of gold answers in each dataset. Similar to \citet{zhang2024tablellm}, we deploy Llama-2-70B-chat~\cite{touvron2023llama} to predict if the concluding sentence answers the question correctly using the prompt in Table~\ref{tab:eval_prompt} in Appendix. Our evaluation results using LLM are checked manually and they align better with human evaluation of the concluding sentence.  
 We report three-class F1 score for \textsc{SciTAB} and accuracy for other datasets.

\noindent\paragraph{In-Context Learning Results} 

The result with GPT-3.5-turbo\footnote{We use \texttt{GPT-3.5-turo-16k-0613} for experiment. We only experiment with WikiTQ, TabFact and FEVEROUS due to limited budgets.} in Table~\ref{tab:results} suggests that \ul{\textit{Plan-then-Reason} framework is effective for answering different types of queries over tables}. Our method surpasses all the baselines without self-consistency on WikiTQ. It also achieves higher accuracy than Dater and Binder, which use 20 responses and 50 responses for self-consistency, respectively. Our method still falls behind Mix SC which predicts answers based on 5 responses with textual reasoning and 5 with program-based reasoning. 
But the self-consistency mechanism leads to a much larger computing cost. 
Our framework also surpasses existing baselines on TabFact except for StructGPT, demonstrating its effectiveness for fact verification tasks. \textit{Plan-then-Reason} can also generalize to tables with sentence context. It increases the accuracy on FEVEROUS by 4.8\%.

\noindent\paragraph{Finetuning Results} 
The experimental results with Llama-2-7B and CodeLlama-7B in Table~\ref{tab:results} show that \ul{our finetuned models generalize to diverse tabular tasks with only 6k training instances}. 
Compared with the backbone model Llama-2-7B, the performance gain of \textsc{ProTrix} on in-domain benchmarks is 34.8\% on WikiTQ and 28.7\% on FEVEROUS. And the performance gain on out-of-domain benchmarks is 21.5\% on average. Comparing the out-of-domain performance with TableLlama, \textsc{ProTrix} surpasses TableLlama by 25.7\% on WikiSQL, 15.8\% on \textsc{SciTAB}, 9.6\% on HybridQA and 11.8\% on TATQA. The overall performance gain on out-of-domain benchmarks demonstrates the planning and reasoning abilities obtained from \texttt{TrixInstruct} is not restricted to in-domain tasks. Our finetuned model \textsc{ProTrix} adaptatively generalizes to queries with different input and output configurations and can even be applied to specific domains such as science and finance. 

\textsc{ProTrix-Coder} still falls behind TableLLM on question answering task with sentence context since TableLLM is finetuned with 8k cases from TATQA. We are surprised to find that \textsc{ProTrix-Coder} surpasses TableLLM on WikiTQ and WikiSQL by 4.9\% and 7.0\%, respectively, even though the training set of TableLLM contains 18k cases from WikiTQ and 28k cases from WikiSQL. Our \texttt{TrixInstruct} only contains 6k training instances in total. These results indicate the effectiveness of finetuning with our \textit{Plan-then-Reason} framework. \textsc{ProTrix-Coder} also surpasses TableLLM on fact verification tasks by 13.8\% on average.

\subsection{Free-form Answer Tasks}
\noindent\paragraph{Baselines} We run GPT-3.5-turbo and TableLlama~\cite{zhang2023tablellama} as our baselines. The prompt for each model is shown in Table~\ref{tab:fetaqa_prompt} in Appendix. We also use the result of finetuning method using T5-large, and human performance from \citet{nan2022fetaqa} as baselines. Notably, the results from \citet{nan2022fetaqa} are evaluated with the original setting where the highlighted cells are provided as input instead of the whole table.

\begin{table}[!t]
\small
\setlength{\tabcolsep}{3pt}
	\centering
	\begin{tabular}{lcccc}
		\toprule
Models & Fluency & Correct & Adequate  &Faithful  \\
 \midrule
T5-large$^{*}$ &94.6 & 54.8 & 50.4 &50.4 \\
  TableLlama & 63.0&67.0 & 55.0&82.0 \\
    \textsc{ProTrix} & 96.0 & 77.0 & 71.0 & 91.0\\
  \midrule
    GPT-3.5-turbo & 99.0 & 83.0 & 85.0 & 96.0\\
  Human performance$^{*}$& 95.0 & 92.4  &95.6 &95.6\\
    \bottomrule
	\end{tabular}
	\caption{Human evaluation results on FetaQA. $^{*}$: results reported by \citet{nan2022fetaqa}.}

	\label{tab:fetaqa_result}
\end{table}

\noindent\paragraph{Evaluation Metrics} Since the response of our model contains step-by-step reasoning over symbolic code and natural language, BLEU~\cite{papineni2002bleu} would underestimate the performance of our model. BLEU also can not be used to evaluate the correctness and faithfulness of the responses. We sample 100 cases from the dataset to perform human evaluation following \citet{nan2022fetaqa}. The evaluation is based on four criteria: (1) \textit{fluency}: if an answer is natural and grammatical; (2) \textit{correctness}: if an answer is correct; (3) \textit{adequacy}: if an answer contains all the information asked by the question; (4) \textit{faithfulness}: if an answer is faithful and grounded to the contents of the table. 

\noindent\paragraph{Results} From Table~\ref{tab:fetaqa_result}, we can observe that \ul{our model exclusively trained on short-form answer tasks can adaptively generalize to give accurate and faithful explanations for complex free-form questions.} Our model achieves a fluency score of 96.0, closely following the human performance at 95.0, indicating its natural and coherent responses.

\textsc{ProTrix} surpasses TableLlama by 33.0\% on \textit{fluency}. TableLlama is observed to lose fluency in some cases where it generates a float number like \textit{2008.0} to answer \textit{what year} or a list of structured \textit{<entity\_name>} which is used to answer entity linking questions from its training set.

Our model achieves \textit{correct} score of 77.0 and \textit{faithful} score of 91.0 which are comparable to GPT-3.5-turbo. Although our model is only trained on short-form answer tasks, the learned planning and reasoning abilities can be utilized to answer complex \textit{how} and \textit{why} questions with accurate and faithful explanations. We present an example of the responses in Table~\ref{tab:fetaqa} in Appendix.

\section{Analysis}
\subsection{Ablation study}
 To better evaluate the effectiveness of finetuning with our \textit{Plan-then-Reason} framework, we experiment with 3 other finetuning frameworks based on \texttt{TrixInstruct}. (1) w/o Planning: We split each instance in \texttt{TrixInstruct} into planning and reasoning parts. We train our model with only the reasoning part of the training instances. This can be considered as distilling the reasoning pattern of TabSQLify~\cite{nahid2024tabsqlify}. (2) w/o Reasoning: Similar to (1), we finetune the model with only the planning part of the training instances. (3) w/o Planning and Reasoning: We finetune the model to generate answers directly. This is similar to the end-to-end QA paradigm.
 \begin{table*}[ht]
\small
\setlength{\tabcolsep}{2pt}
	\centering
	\begin{tabular}{lccccccc}
		\toprule
 Models & WikiTQ & WikiSQL & Tabfact& \textsc{SciTAB} & FEVEROUS & HybridQA & TATQA \\
 \midrule
\textsc{ProTrix} & \textbf{53.8}&\textbf{65.7} &\textbf{73.4}&\textbf{45.0} &\textbf{75.6} & \textbf{42.9}&50.1 \\
\quad w/o Planning& 51.0 & 63.9 & 66.4 & 31.8& 66.8 & 41.3 & \textbf{50.4} \\
\quad w/o Reasoning& 41.4 & 54.3 & 65.4 & 33.4 & 70.4 & 36.3 & 39.8\\
\quad w/o Planning and Reasoning& 39.5 & 47.8  & 59.0 & 29.4& 64.8& 29.4 &35.3\\
    \bottomrule
	\end{tabular}
	\caption{Ablation study}
	\label{tab:ablation_study}
\end{table*}

The result of the ablation study is presented in Table~\ref{tab:ablation_study}. 
\ul{Both planning and reasoning contribute significantly to the overall performance and generalizability of our model.} 
Excluding planning or reasoning would cause the average performance to decrease by 5.0\% or 9.4\%, respectively. 
In w/o planning setting, the performance on \textsc{SciTAB} and FEVEROUS drops significantly by 13.2\% and 8.8\%, respectively. It suggests the importance of planning ability in utilizing commonsense knowledge and decomposing the query into reasoning chains over tables and sentences. 
The w/o planning and reasoning setting is similar to previous methods that train the model to answer queries directly~\cite{xie2022unifiedskg, zhang2023tablellama, zhuang2024structlm}. The performance of in-domain and out-of-domain benchmarks drops by 14.1\% and 15.2\% on average, emphasizing the effectiveness of the \textit{Plan-then-Reason} framework in promoting generalizability across tabular tasks.

\subsection{Prompting Efficiency Analysis}
\label{app:efficiency_analysis}

In Table~\ref{tab:prompt_method_comparison}, we compare the number of annotated in-context demonstrations and average API calls of different prompting methods on WikiTQ. It shows that \ul{our Plan-then-Reason framework can effectively reach the final answer with less in-context demonstrations and API calls}. Binder~\cite{cheng2022binding}, Dater~\cite{ye2023large}, Chain-of-Table~\cite{wang2024chain} and TabSQLify~\cite{nahid2024tabsqlify} require more than 10 annotated in-context examples for inference while our \textit{Plan-then-Reason} framework only needs one in-context demonstration. Binder, Dater and MIX SC predict the answer with the self-consistency mechanism which needs at least 10 API calls. Chain-of-Table follows the style of ReAct~\cite{yao2023react} with averagely 10.1 API calls to reach the final answer, while \textit{Plan-then-Reason} only requires 2 calls on average.

\begin{table}[h]
\small
\setlength{\tabcolsep}{3pt}
	\begin{tabular}{lcc}
    \toprule
   Method &  Annotated Examples & Average API calls\\
   \midrule
   Binder & 14 & 50\\
   Dater & 17 & 100\\
   Chain-of-Table & 29 & 10.1\\
    TabSQLify & 12 & 2\\
    MIX SC & 0 &  10\\
    Plan-then-Reason & 1 & 2 \\

    \bottomrule
	\end{tabular}
        \caption{Comparison of the number of annotated examples and average API calls for WikiTQ.}
	\label{tab:prompt_method_comparison}
\end{table}

\subsection{Error Analysis}
\label{sec:error_analysis}
We perform error analysis on the results of \textit{Plan-then-Reason} with GPT-3.5-turbo and \textsc{ProTrix} on WikiTQ and FEVEROUS to analyze the potential weakness of our model.
We categorize the error cases into the following types: (1) Planning error: the response fails to generate a correct plan to answer the question, (2) SQL error: The response generates SQL containing columns that do not exist at all or the SQL simply can not be executed. (3) Execution error: the execution result given by GPT-4 is wrong. (4) Reasoning error: the model achieves a correct answer with wrong reasoning. (5) False Negative: the gold answer is wrong or the executor misjudges the result. We show examples for each error type in Table~\ref{tab:error_examples} in Appendix. We also perform qualitative analysis shown in Appendix~\ref{app:case_study}.

\begin{table}[h]
\centering
\small
\setlength{\tabcolsep}{3pt}

\begin{tabular}{lcccc}
\hline
\multirow{2}{*}{Error Type} & \multicolumn{2}{c}{WikiTQ} & \multicolumn{2}{c}{FEVEROUS} \\  
                           & \textsc{ProTrix} & GPT-3.5 & \textsc{ProTrix} & GPT-3.5\\ \hline
Planning Error             & 22\%    & 18\%          & 36\%    & 46\%          \\ 
SQL Error                  & 32\%    & 26\%          & 22\%    & 12\%          \\ 
Execution Error            & 30\%    & 28\%          & 10\%    & 8\%           \\ 
Reasoning Error            & 12\%    & 14\%          & 26\%    & 26\%          \\ 
False Negative             & 4\%     & 14\%          & 6\%     & 8\%           \\ \hline
\end{tabular}
\caption{Error analysis of \textsc{ProTrix} and GPT-3.5-turbo on WikiTQ and FEVEROUS.}
\label{tab:error_analysis}
\end{table}

Despite the effectiveness of our framework in both in-context learning and finetuning methods, we identify two primary sources of errors.
The first error source is \ul{generating SQL queries for complex table content or program structures.}
The main challenge for WikiTQ is to write the correct SQL to perform complex operations on the table. The model sometimes generates SQLs with invalid program structures for questions that need multi-step table operations such as comparing the results of two maximum values under different conditions. The model also struggles to write SQL to process complex context such as isolating country names from table cells as \textit{John (ESP)} where the data is embedded within the text. We think it is due to the limited coding ability in generating SQLs for multi-step operations or complex cell content. We plan to enhance the specific coding ability for SQL of ProTrix in future work.

The second main error source is \ul{decomposing complex queries into sub-questions.} FEVEROUS challenges the model to decompose a claim into several sub-questions, digest the relevant table or sentences to answer each sub-question, and then combine the results to form the final answer. In some cases, \textsc{ProTrix} is observed to inaccurately assign sub-questions that are directly related to the sentences to be answered using tables, or it may overlook certain sub-questions during the decomposition of the original claim. We suspect that the error source of missing sub-questions is that the model can not retain all the sub-questions from the claim during planning due to its autoregressive generation nature. We plan to implement a memory mechanism of sub-questions to address this issue.

\section{Related Work}
\noindent\paragraph{Prompting Methods for LLMs} 
Large language models can be guided to solve tasks in a step-by-step manner~\cite{wei2022chain, hao-etal-2023-reasoning}. \citet{chen2023large} first utilizes Chain-of-Thought~\cite{wei2022chain} to enhance the reasoning of LLMs on tabular tasks and points out that textual reasoning can not generalize to large tables directly. Researchers prompt the model to select relevant rows and columns as one step in the chain of reasoning to enable LLM to focus on the following reasoning step~\cite{jiang-etal-2023-structgpt, ye2023large, wang2024chain}. \citet{chen2022program} proposes Program-of-Thoughts (PoT) that answers a question in programming language. Compared with textual reasoning, program-based reasoning is executed by a code interpreter, achieving high-precision reasoning in complex tabular or mathematical questions. 
Binder~\cite{cheng2022binding} binds LLMs as API calls within a Python or SQL program to address the program-unsolvable aspect of the queries.  \citet{liu2023rethinking} proposes mix self-consistency that combines the potential of both textual and program-based reasoning. Researchers have also attempted ReAct~\cite{yao2023react} style prompting for tabular tasks. 
ReAcTable~\cite{zhang2023reactable} reasons with SQL or Python tools in multiple turns. Chain-of-Table~\cite{wang2024chain} formats table reasoning as specific table operations. ReAct-style prompting mainly focuses on the next acting step based on feedback from the last action without considering the whole reasoning chain. Our \textit{Plan-then-Reason} framework can plan the whole reasoning chain before acting and answer questions more efficiently and accurately (see efficiency analysis in Appendix~\ref{app:efficiency_analysis}) while blending the advantages of textual and program-based reasoning methods.

\noindent\paragraph{Finetuned Models} 
Various pre-trained models are proposed for tabular tasks~\cite{yin2020tabert,wang2021tuta,iida2021tabbie,deng2022turl,yang2022tableformer,jiang2022omnitab,liu2021tapex}. But they often are limited to one specific downstream finetuning task. As for models with generalizability,
\citet{liu2023zero} mix symbolic SQL execution task with FLAN task to further finetune FLAN-T5 to improve zero-shot tabular question answering performance. 
\citet{li2023table} finetunes models with a large synthesized dataset of table manipulation and data augmentation to serve as a table-foundation model that understands table structures.
 TableLlama~\cite{zhang2023tablellama} and StructLM~\cite{zhuang2024structlm} collect an instructing tuning set that 
 covers diverse tables and tasks and finetune Llama to obtain a generalist model without table pretraining. Compared with existing generalist models that are expected to generate answers directly, \textsc{ProTrix} is interpretable by generating accurate and faithful explanations. 
 
 TaCo \cite{zheng-etal-2023-chain} is finetuned with step-by-step solutions of math problems over tabular data. However, it is only limited to mathematical table reasoning and can not generalize to other types of tabular tasks.  \citet{zhang2024tablellm} uses textual or program-based reasoning for different in-domain benchmarks without planning the reasoning method based on the query and the context. 
 \textsc{ProTrix} can plan over tables with sentence context and assign each step to textual or program-based reasoning to reach the final answer.

\section{Conclusions}
In this paper, we propose a novel \textit{Plan-then-Reason} framework to answer user queries over tables with sentence context, which analyzes the commonsense and concepts in the query and plans the reasoning steps over programs and natural languages. We construct an instruction tuning set \texttt{TrixInstruct} to finetune models to obtain such planning and reasoning abilities with only 6k examples. Experiments show that our resulting models \textsc{ProTrix} and \textsc{ProTrix-Coder} can generalize to unseen tabular tasks with sentence context and produce accurate and faithful explanations.
Our work highlights the required abilities for generalist models over tabular tasks with sentence context, and paves the way for future research directions.

\section*{Limitations}
The instances in \texttt{TrixInstruct} only contain relational tables. It currently does not contain complex tables with hierarchical headers~\cite{cheng2022hitab}. 
And  \texttt{TrixInstruct} is restricted to queries over one table. It can not be directly applied to tabular tasks over multiple tables or retrieved top $k$ tables. We plan to extend \texttt{TrixInstruct} to cover more realistic scenarios in future work.

We find it hard to control the rule or grammar of the generated answer by open-source models and exact match often fails to evaluate the performance properly due to format issues, especially for out-of-domain benchmarks. We tried several ways as evaluation method and choose LLMs as evaluator. Since we use an exact match for GPT-3.5-turbo results, the performance between open-source and closed-source models can not be directly compared.

\section*{Acknowledgement}
This work is supported in part by NSFC (62161160339) and Beijing Science and Technology Program (Z231100007423011). We would like to thank the anonymous reviewers
for their comments and suggestions. We also thank Rami Aly and Md Mahadi Hasan Nahid for the helpful discussions. 
For any correspondence, please contact Yansong Feng.

\bibliography{custom}

\appendix

\section{Implementation Details}
\label{app:detail}
We fully finetune Llama-2 7B~\cite{touvron2023llama} and CodeLlama-7B~\cite{roziere2023code} with our instruction tuning set with the context of length 4096. We set the learning rate as 5e-6 and the batch size as 32. The training process uses a cosine scheduler with a 3\% period for 3 epochs. We utilize DeepSpeed training with ZeRO-3 stage~\cite{rasley2020deepspeed}. Our model is trained with 4 NVIDIA A40 GPUs (48GB) and the whole training process takes about 5 hours. 

During inference for \textsc{ProTrix} and GPT-3.5-turbo, we set the output length as 1024, temperature as 0 and truncate large tables to fit in context length. Then we prompt the model to generate a response for the query, if there is a SQL in the response, we replace the execution result with an output of the actual SQL execution tool and ask the model to generate the rest of the response. If the SQL can not be executed, we fall back to the execution result the model generates. 

For finetuned SOTA methods in Table~\ref{tab:results}, we report the performance of OmniTab~\cite{jiang2022omnitab} for WikiTQ, TAPEX~\cite{liu2021tapex} for WikiSQL, PASTA~\cite{gu2022pasta} for TabFact, S$^{3}$HQA~\cite{lei2023s} for HybridQA and APOLLO~\cite{sun2022apollo} for TATQA.  For FEVEROUS, we run DCUF~\cite{hu2022dual} on our training and development set of FEVEROUS and obtain an accuracy of 75.9\%.  Notice that S$^{3}$HQA uses a more precise sentence retriever compared to ours and DCUF leverages an additional retriever to select 25 table cells as input.

\section{Training Dataset Analysis}
\label{app:train_analysis}
Our instruction tuning dataset is extracted from GPT-4 responses. We filter out the responses that have inconsistent final answers with the original dataset annotations,
but the reasoning process of the responses in the training set has not been fully validated. 
We perform a quality analysis of our training dataset following the error types defined in \S~\ref{sec:error_analysis}
We sample 50 cases with sentence context and 50 cases without sentence context from \texttt{TrixInstruct} to perform manual evaluation.

\begin{table}[ht]
\small
\setlength{\tabcolsep}{5pt}
	\centering
	\begin{tabular}{lcccc}
    \toprule
    & w/o Sentence  & w/ Sentence  & Overall\\
    \midrule
    Planning Error &  4\%& 10\%& 7\%\\
    SQL Error & 10\%&8\%& 9\%&\\
    Execution Error &2\% & 2\% & 2\%\\
    Reasoning Error & 2\% & 6\%& 4\%\\
    Accurate & 82\% & 74\% & 78\%\\
    \bottomrule
	\end{tabular}
        \caption{Quality analysis of \texttt{TrixInstruct}. w/(w/o) Sentence: subset of queries over tables with(without) sentence context.}
	\label{tab:training_analysis}
\end{table}

The analysis of the instruction tuning set reveals distinct error patterns in responses generated for queries over tables with or without sentence context. In the subset without sentence context, the most prevalent error type is SQL errors, indicating issues with SQL query generation for some complex questions, including referencing non-existent columns or generating unexecutable queries. As for the subset with sentence context, the most prevalent error type is planning errors. The response fails to decompose the claim into sub-claims or generates wrong plans to query the tables. Notably, both subsets showcase minimal execution errors, suggesting the proficiency of GPT-4 in providing accurate execution results. Reasoning errors are more common in the subset with sentence error with an error rate of 6\%. 
We observe that the reasoning process generated by GPT-4 can not always follow the reasoning chains designed during planning.
Overall, the manual analysis of \texttt{TrixInstruct} indicates a combined 7\% planning error rate, a 9\% SQL error rate, a 2\% execution error rate and a 4\% reasoning error rate.  78\% of instances reach correct answers with accurate planning and reasoning process.

We have checked all the instances in \texttt{TrixInstruct} and make sure they do not contain any private information or offensive content.

\section{Analysis of Breakdown Performance}

\subsection{Program-Unsolvable Queries}
To analyze the performance on queries that need commonsense knowledge or textual reasoning. We decompose the original development set of WikiTQ
into program-solvable and program-unsolvable subsets following \citet{shi2020squall}. We compare the performance of \textsc{ProTrix} and \textsc{ProTrix-Coder} with Binder~\cite{cheng2022binding},UnifiedSKG~\cite{xie2022unifiedskg}, TAPEX~\cite{liu2021tapex} and TaCube~\cite{zhou-etal-2022-tacube}. Notably, our models are only trained with less than 3k examples from WikiTQ while TAPEX and TaCube are trained on the original training set which contains over 11k examples. UnifiedSKG is trained on 21 tasks involving WikiTQ. Binder prompts Codex to write code with LLMs as APIs. We do not compare with TableLlama since it is not trained on WikiTQ.

From Table~\ref{tab:wikita_decompose}, we can observe that \textsc{ProTrix-Coder} achieves the highest accuracy on program-unsolvable queries compared with finetuned methods. 
It suggests \texttt{TrixInstruct} can teach a model to understand commonsense and general concepts in the query and adaptatively plan to reason with programs or languages. \textsc{ProTrix-Coder} still falls behind TAPEX and TaCube on the program-solvable subset. But these models require table pretraining which is computationally expensive. \textsc{ProTrix-Coder} surpasses the previous generalist model by 1.5\% and 4.7\% on program unsolvable and solvable subsets, indicating the effectiveness of the proposed \textit{Plan-then-Reason} framework.

\begin{table}[ht]
\small
\setlength{\tabcolsep}{3pt}
	\begin{tabular}{lccc}
    \toprule
   Models & P-Unsolvable & P-Solvable & Overall\\
    \midrule
    \textit{Closed-source Models}& & & \\
    Codex & 40.3 & 53.4 & 50.5\\
    Binder & 41.3 & 71.8 & 65.0\\
    \midrule
    \textit{Finetuning Methods} & & & \\
    UnifiedSKG& 37.6 & 56.0 & 51.9\\
    TAPEX$^{*}$ & 33.6 & 68.0 & 60.4\\
    TaCube$^{*}$&34.9 &68.5 & 61.1\\
    \midrule
    \textsc{ProTrix} &35.0 &59.1 &53.8 \\
    \textsc{ProTrix-Coder} & 38.9&60.7 &55.7\\

    \bottomrule
	\end{tabular}
        \caption{Breakdown performance on the development set of WikiTQ. P-(un)solvable: program-(un)solvable subset. $^{*}$: with table pretraining.}
	\label{tab:wikita_decompose}
\end{table}
\subsection{Combining Tables and Sentences}
We break down the performance on FEVEROUS into subsets following \citet{aly2021feverous}. We choose subsets that are related to the planning and reasoning abilities to analyze our model as shown in Table~\ref{tab:feverous_decompose}.

We use GPT-3.5-turbo and DCUF~\cite{hu2022dual} as baselines.  
Notably, our reproduction of DCUF leverages an additional module~\cite{wu2023enhancing} to select top 25 cells from the table to control input context length. GPT-3.5-turbo and our models use the whole table as input.

From Table~\ref{tab:feverous_decompose}, we can observe that  \textsc{ProTrix} has comparable performance with GPT-3.5-turbo and DCUF on combining tables and texts and multi-hop reasoning. It suggests that our model can learn to plan the reasoning steps and assign them to programs or languages by training on \texttt{TrixInstruct}. \textsc{ProTrix} surpasses GPT-3.5-turbo and DCUF by 25.5\% and 5.4\%, respectively, on the numerical reasoning subset. It underscores that 
symbolic programming can achieve high-precision performance.

\begin{table}[th]
\small
\setlength{\tabcolsep}{3pt}
	\begin{tabular}{lccc}
    \toprule
   Models & Table+Text & Multi-hop & Numerical\\
    \midrule
    \textit{Closed-source Models}& & & \\
    GPT-3.5-turbo& 81.3&79.2 & 48.6 \\
    \midrule
    \textit{Finetuning Methods} & & & \\
    DCUF$^{\dagger}$ & 83.4 & 77.8 & 68.7\\
    \midrule
    \textsc{ProTrix} & 81.8& 73.9&74.1 \\
    \textsc{ProTrix-Coder} & 78.1 & 68.8 &73.1\\

    \bottomrule
	\end{tabular}
        \caption{Breakdown performance on our development set of FEVEROUS. Table+Text: combining tables and texts. Multi-hop: multi-hop reasoning. Numerical: Numerical reasoning. $^\dagger$: select top 25 cells from the table as input following \citet{wu2023enhancing}.}
	\label{tab:feverous_decompose}
\end{table}

\section{More Results}
In Table~\ref{tab:detailed_results}, we compare the results with more closed-source models. Even GPT-3.5-turbo falls behind Codex with same prompting methods, our \textit{Plan-then-reason} is comparable to results of Binder, ReAcTable and Dater using Codex. We also experiment with prompting methods on 7B models to understand the effectiveness of our model, especially for out-of-domain benchmarks. We run StructGPT~\cite{jiang-etal-2023-structgpt}, Plan-and-Solve~\cite{wang2023plan}, and Chain-of-Thought~\cite{wei2022chain} on Llama-2-7B-chat. We also run PAL~\cite{gao2023pal}, ReAct~\cite{yao2023react} and ARC~\cite{zhang2023crt} on CodeLlama-7B-Instruct. Our \textsc{ProTrix} model family surpasses all the prompting methods with 7B models. 

We also experiment with the latest Llama-3-8B model. The performance increases by 9.5\% on average. It shows that our dataset can be used to finetune larger and more powerful base models to obtain more accurate answers.

\section{Case Study}
\label{app:case_study}
We conduct case studies to further demonstrate the planning and reasoning ability of \textsc{ProTrix}. Table~\ref{tab:wikitab_case} demonstrates the planning ability that digests the general concept in user query and fills the gap between the question \textit{who was the top chef} and the table of competition result. As for the example in Table~\ref{tab:tabfact_case}, \textsc{ProTrix} deploys SQL to extract relevant information from the table and reach the final answer based on textual reasoning. \textsc{ProTrix} also generalizes to queries with insufficient context. We can observe from Table~\ref{tab:scitab_case}, that the model refuses to write SQL when the table does not contain the necessary information to verify the claim. The example in Table~\ref{tab:feverous_case} shows the ability to decompose query and assign each step to program-based or textual reasoning and the example in Table~\ref{tab:hybridqa_case} suggests the ability to plan reasoning steps based on sentence context.

We also demonstrate some examples that \textsc{ProTrix} fails to answer correctly. Consider the example in Table~\ref{tab:error_tat}, \textsc{ProTrix} successfully extracts relevant cells from the table but fails to compute the final answer using textual reasoning. It indicates that there is still room for improvement on numerical reasoning.
The example in Table~\ref{tab:error_feverous} is a fact verification query. The table does not have hierarchical headers but has an intrinsic hierarchical structure. The last \textit{Total} row should not be considered as a \textit{HDI} and our model fails to identify the intrinsic hierarchical structure. We plan to include tables with hierarchical headers and structure in our instruction tuning dataset in future work.

\begin{table*}
\centering
\small
\setlength{\tabcolsep}{5pt}
\begin{tabular}{p{0.3\columnwidth}<{\centering}p{0.6\columnwidth}p{0.8\columnwidth}}
\toprule
\textbf{Error Type} & \textbf{Claim} & \textbf{Response}\\
\midrule
Planning Error &McCord Museum is a history museum containing 1,440,000 objects, images and manuscripts. & - To verify the claim, we need to \red{check the collection size of the McCord Museum}. This information can be found in the table context under the "Collection size" column.
- We do not need to use the sentence context for this claim as the table provides the exact information needed.\\
\midrule
SQL Error & Georgia Southern did not make Athlon's list of preseason top 25 of 2011 NCAA Division I FCS football rankings. (column 0 is a column for rankings) & SELECT column0 FROM w WHERE \red{column0 = 'Georgia Southern'}
\\
\midrule
Execution Error & In 1949, the Chicago Cardinals season played twelve matches, resulting to ten losses and two tied matches.& SELECT COUNT(*) as Losses FROM w WHERE Result LIKE 'L\%' \red{Result: 4}\\
\midrule
Reasoning Error &In Southern Peninsular Malaysian Hokkien phonology, there are three front vowels, two central vowels and two back vowels.
 & - The claim states that in Southern Peninsular Malaysian Hokkien phonology, there are three front vowels  two central vowels, and \red{four back vowels}.
...
Therefore, the answer is REFUTES.\\
\bottomrule
\end{tabular}
\caption{Types of errors in \textit{Plan-then-Reason} framework. For each response, we only demonstrate the part of the response where the error first occurs. The error is highlighted in red. In planning error, the model response fails to list all the information that needs to be checked apart from collection size. In SQL error, the SQL can not be used to extract the ranking of Georgia Southern. As for reasoning error, the model fails to summarize the whole planning and reasoning because of the hallucination of the original claim.}
\label{tab:error_examples}
\end{table*}

\begin{table*}[th]
\small
\setlength{\tabcolsep}{4pt}
	\centering
	\begin{tabular}{lccccccc}
		\toprule
& WikiTQ & WikiSQL & TabFact & \textsc{SciTAB} &FEVEROUS & HybridQA &TATQA\\
 \midrule
 \textit{GPT-4} &  &  &  &  & & &\\
 \quad End-to-End QA & 72.9 & 75.8 & 71.5 & 57.1& 71.0& 64.1 & 80.8\\
 \midrule
 \textit{PaLM2} &  &  &  &  & & &\\
\quad End-to-end QA & 60.5 & & 77.9& - & - & - & - \\
\quad Chain-of-Table  & 67.3 & - & 86.6 & - & - & - & - \\
\midrule
\textit{Codex} &  &  &  &  & & &\\
\quad Binder & 64.6 & & 85.1 & & & & \\
\quad ReAcTable  & 65.8 & - & 83.1 & - & - & - & - \\
\quad Dater  & 65.9 & - &85.6 &-& - & - &-\\
\quad SEER & - & - & - & - & - & - & 73.6 \\
\midrule
 \textit{GPT-3.5-turbo} &  &  &  &  & & &\\
 \quad End-to-End QA & 51.8& 55.0 & 68.8&45.3 & 61.0& 55.1 & 59.1\\
\quad ReAcTable  & 52.4 & - & 73.1 & - & - & - & - \\
\quad Dater$^{*}$ & 52.8 & - & 78.0 & - & - & - & - \\
\quad Binder$^{*}$  & 56.7 & - &79.2 &-& - & - &-\\
\quad StructGPT  & 57.0 & 64.6 & \textbf{87.3} & - & - & - & - \\
\quad Chain-of-Table & 59.9 & - & 80.2 & - & - & - & - \\
\quad TabSQLify & 64.7 & 76.7 & 79.5 &  - & - & - & - \\
\quad Mix SC$^{*}$ & 73.1  & - & - &  - & - & - & - \\
\quad Plan-then-Reason$^{\ddagger}$(Ours)& 60.5& -& 79.6 &- &53.8  & -& -\\
\quad Plan-then-Reason (Ours)& 65.2& -& 83.5 &- & 65.8 &- & -\\
 \midrule
\textit{Finetuned SOTA} & ~~63.3$^{\dagger}$ & ~~89.2$^{\dagger}$ & ~~90.8 $^{\dagger}$& - & ~~75.9$^{\dagger}$ & ~~61.0$^{\dagger}$&~~74.5$^{\dagger}$ \\
 \midrule
 \textit{Llama-2-7B} & & & & &  \\
   \quad End-to-End QA & 21.4& 17.4 &48.6 &27.2& 47.1 & 27.6 & 28.7\\
      \quad Llama-2-7B-chat StructGPT & 21.2 & 23.1 & 38.9 & 30.5 & 16.1 & 27.8 & 21.3\\
\quad Llama-2-7B-chat PS  & 26.1 & 25.1 & 31.7 & 31.4& 39.4 &24.6&36.4\\
\quad Llama-2-7B-chat CoT & 33.8 & 28.4 & 49.8 &36.6 & 44.8 & 24.9 & 35.8\\
   \quad TableLlama & 31.6 & 41.7 & ~~82.6$^{\dagger}$&29.2 & 56.8 & 33.3&38.3 \\
    \quad \textsc{ProTrix}(Ours) & ~~56.2$^{\dagger}$ & 67.4 & 71.6 & 45.0& ~~75.6$^{\dagger}$ & 42.9&50.1\\
    \midrule
  \textit{CodeLlama-7B} & & & & &  \\
    \quad End-to-End QA & 13.1 & 17.3 & 49.5&37.1& 43.0 &28.5 &28.4\\
\quad CodeLlama-7B-Instruct PAL  & 24.5 & 11.5 & 33.1 & 30.2 & 27.2 & 6.1 &  11.1\\
\quad CodeLlama-7B-Instruct ReAct  & 34.2 & 38.4 & 52.6 & 15.3 & 43.2 & 19.1 & 34.9\\
\quad CodeLlama-7B-Instruct ARC  & 35.8& 39.9 & 54.6 & 29.5 & 49.2 &23.7 &28.2\\
\quad TableLLM & ~~52.9$^{\dagger}$& ~~65.3$^{\dagger}$& 57.1&24.7 & 60.0 &\textbf{53.7} & ~~\textbf{70.3}$^{\dagger}$\\
    \quad \textsc{ProTrix-Coder} (Ours) & ~~57.8$^{\dagger}$&  72.3 & 70.6 &41.2& ~~71.4$^{\dagger}$ & 45.1  & 52.2\\
    \midrule
  \textit{Llama-3-8B} & & & & &  \\
    \quad Llama-3 & 29.3& 31.8 & 54.8 & 28.6 & 48.3 &32.8 & 45.4 \\
 \quad \textsc{Llama-3-ProTrix } (Ours)& ~~67.0$^{\dagger}$& 75.7& 79.4 & 41.3& ~~84.5$^{\dagger}$& 57.3 & 70.3\\

    \bottomrule
	\end{tabular}
	\caption{Experimental results on short-form question answering and fact verification tasks. $^{\dagger}$ The model is trained on this evaluation benchmark. $^{*}$: with self consistency. We use \texttt{GPT-4-1106-preview} and \texttt{GPT-3.5-turbo-16k-0613} in our experiment. Plan-then-Reason$^{\ddagger}$: w/o fallback mechanism that use LLM to predict the result of code that fails to execute.}

	\label{tab:detailed_results}
\end{table*}

\begin{table*}
\centering
\small
\setlength{\tabcolsep}{5pt}
\begin{tabular}{p{2.0\columnwidth}r}
\toprule
This is a table question answering task. You will answer the question based on the given context. 
\\ \\
\#\# \textbf{Question}\\
How many medals were won by Kenya?\\ \\

\#\# \textbf{Table Context}\\
| Rank | Name              | Nationality    | Time    |\\
| 1    | Brahim Boulami    | Morocco        | 8:17.73 |\\
| 2    | Reuben Kosgei     | Kenya          | 8:18.63 |\\
| 3    | Stephen Cherono   | Kenya          | 8:19.98 |\\
| 4    | Bouabdellah Tahri | France         | 8:20.25 |\\
| 5    | Tim Broe          | United States  | 8:20.75 |\\
| 6    | Luis Miguel Martín| Spain          | 8:24.03 |\\
| 7    | Raymond Yator     | Kenya          | 8:27.19 |\\
| 8    | Thomas Chorny     | United States  | 9:24.26 |\\ \\

\#\# \textbf{Task}\\
You will answer the question based on the given context.You should reach a short-form answer after reasoning.
You are asked to answer the question in three steps.\\
1. Analyze the question and the given context. Make up a plan to answer the question.\\
2. Write one or more SQL to query the table for necessary information and output expected execution result.\\
3. Reason step-by-step to reach the final answer.\\ \\

\#\# \textbf{Answer}\\
\\
Let's answer in three steps.\\
\textbf{1. Plan for answering the question}\\
- The question focus on the medals won by Kenya. According to common sense, only top 3 players can win medals. \\
- So we write a SQL to select the top three players. Then we count the numbers of medals won by Kenya based on the output of SQL.
\\
2.\textbf{ Write SQL and execute SQL} \\
SQL:\\
SELECT Name, Nationality FROM Table WHERE Rank <= 3\\
\\
Executed result:\\

| Name            | Nationality |\\
| Reuben Kosgei   | Kenya       |\\
| Stephen Cherono | Kenya       |\\ \\

3. \textbf{Step-by-Step Reasoning}:\\
- The question is asking about the number of medals won by Kenya.

- Based on the execution result of SQL query, only two of the players that win medals are from Kenya.

- Therefore, the total number of medals won by Kenya is 2.
\\ \\
The final answer is \textbf{2}.\\\\
\#\# \textbf{Question}\\
\textit{Question}\\ \\

\#\# \textbf{Table Context}\\
\textit{Table}\\
\\
\#\# \textbf{Task}\\
You will answer the question based on the given context.You should reach a short-form answer after reasoning.
You are asked to answer the question in three steps.\\
1. Analyze the question and the given context. Make up a plan to answer the question.\\
2. Write one or more SQL to query the table for necessary information and output expected execution result.\\
3. Reason step-by-step to reach the final answer.\\ \\

\#\# \textbf{Answer}\\
\bottomrule
\end{tabular}
\caption{Prompt for generating responses for queries from WikiTQ. GPT-3.5-turbo and GPT-4 generates responses following this example.}
\label{tab:wikitab_distill}
\end{table*}

\begin{table*}
\centering
\small
\setlength{\tabcolsep}{5pt}
\begin{tabular}{p{2.0\columnwidth}r}
\toprule
This is a fact verification task. You are asked to check the veracity of the claim. Both table and sentence context are provided but you are not required to use both of them. You can use either of them or both of them.
\\ \\
\#\# \textbf{Claim}\\
Sony Dwi Kuncoro, born on July 7, 1984, was the champion of the Men's singles event of the 2009 Indonesia National Badminton Championship with a score of 21-17, 22-20. 
\\ \\
\#\# \textbf{Table Context}\\
Page Title: 2009 Indonesia National Badminton Championship\\
Caption:  \\
Category | Winners | Runners-up | Score\\
Men's singles | Sony Dwi Kuncoro | Andre Kurniawan Tedjono | 21-17, 22-20\\
Women's singles | Maria Kristin Yulianti | Maria Febe Kusumastuti | 21-17, 17-21, 23-21\\
Men's doubles | Hendra Setiawan | Rian Sukmawan , Yonathan Suryatama Dasuki | 21-9, 21-13\\
Women's doubles | Meiliana Jauhari , Greysia Polii | Liliyana Natsir , Shendy Puspa Irawati | 21-19, 18-21, 21-17\\
Mixed doubles | Tontowi Ahmad , Greysia Polii | Muhammad Rijal , Debby Susanto | 21-11, 21-19".\\ \\

\#\# \textbf{Sentence Context}\\
Sony Dwi Kuncoro (1): Sony Dwi Kuncoro (born 7 July 1984) is a badminton singles player from Indonesia.\\
Sony Dwi Kuncore (2): He reached a career high as world number 3 on 7 October 2004.\\
Sony Dwi Kuncore (3): In 2009, his best performance in Superseries was semifinalist in Indonesia Open, beaten by Taufik Hidayat, 17–21, 14–21 and in Denmark Open beaten by Marc Zwiebler of Germany in three tough games.\\
\\
\#\# \textbf{Task}\\
You will predict the claim's label based on the given claim and context. There are 3 possible labels: \textit{SUPPORTS}, \textit{REFUTES}, and \textit{NOT ENOUGH INFO}. \textit{SUPPORTS} means the claim is supported by the context. \textit{REFUTES} means the claim is refuted by the context. \textit{NOT ENOUGH INFO} means the context does not provide enough information to the support or refute the claim.
You are asked to check the claim in three steps.\\
1. Analyze the question and the given context. Make up a plan to verify the claim.\\
2. Write one or more SQL to query the table for necessary information and output expected execution result.\\
3. Reason step-by-step to reach the final answer.\\ \\

\#\# \textbf{Answer}:\\
Let's check the claim in 3 steps.\\\\
\textbf{1. Plan for verify the claim}\\
- Based on the claim and the given context. The information for winner of Men's singles can be obtained from the table. We could write a SQL to query the name of the winner and the score of the final for Men's singles.\\
- The claim also states that Sony Dwi Kuncoro is born on July 7, 1984. We will use the sentences to check this part of claim. \\ \\

\textbf{2. Write SQL and execute SQL}\\
SQL:\\
SELECT Winners FROM Table WHERE Category = Men's singles\\
\\
Executed result:\\

Category | Winners  | Score\\
Men's singles | Sony Dwi Kuncoro | 21-17, 22-20 \\ \\

3. \textbf{Step-by-step Reasoning}\\
- The claim states Sony Dwi Kuncoro, born on July 7, 1984, was the champion of the Men's singles event of the 2009 Indonesia National Badminton Championship with a score of 21-17, 22-20. \\

- The SQL query result supports the claim. Sony Dwi Kuncoro was the champion of the Men's singles event of the 2009 Indonesia National Badminton Championship with a score of 21-17, 22-20.\\

- The sentence evidence Sony Dwi Kuncoro (1) supports that he was indeed born on July 7, 1984.\\
\\
Therefore, the answer is \textbf{SUPPORTS}\\\\
\textbf{\#\# Claim}\\
...\\
\textbf{\#\# Answer}\\
\bottomrule
\end{tabular}
\caption{Prompt for generating responses for queries from FEVEROUS. GPT-3.5-turbo and GPT-4 generates responses following this example.}
\label{tab:feverous_distill}
\end{table*}

\begin{table*}
\centering
\small
\setlength{\tabcolsep}{5pt}
\begin{tabular}{p{2.0\columnwidth}r}
\toprule
This is a table fact verification task. You will predict whether the statement is true or false based on the given context.
\\ \\
\#\# \textbf{Claim}\\
Kenya won 3 medals in the 3000 meters event at the 2001 Goodwill Games.\\ \\

\#\# \textbf{Table Context}\\
| Rank | Name              | Nationality    | Time    |\\
| 1    | Brahim Boulami    | Morocco        | 8:17.73 |\\
| 2    | Reuben Kosgei     | Kenya          | 8:18.63 |\\
| 3    | Stephen Cherono   | Kenya          | 8:19.98 |\\
| 4    | Bouabdellah Tahri | France         | 8:20.25 |\\
| 5    | Tim Broe          | United States  | 8:20.75 |\\
| 6    | Luis Miguel Martín| Spain          | 8:24.03 |\\
| 7    | Raymond Yator     | Kenya          | 8:27.19 |\\
| 8    | Thomas Chorny     | United States  | 9:24.26 |\\ \\

\#\# \textbf{Task}\\
You will answer the question based on the given context.You should reach a short-form answer after reasoning.
You are asked to answer the question in three steps.\\
1. Analyze the question and the given context. Make up a plan to answer the question.\\
2. Write one or more SQL to query the table for necessary information and output expected execution result.\\
3. Reason step-by-step to reach the final answer.\\ \\

\#\# \textbf{Answer}\\
Let's answer in three steps.\\
\textbf{1. Plan for answering the question}\\
- The question focus on the medals won by Kenya. According to common sense, only top 3 players can win medals. \\
- So we write a SQL to select the top three players. Then we count the numbers of medals won by Kenya based on the output of SQL.
\\
2.\textbf{ Write SQL and execute SQL} \\
SQL:\\
SELECT Name, Nationality FROM Table WHERE Rank <= 3\\
\\
Executed result:\\

| Name            | Nationality |\\
| Reuben Kosgei   | Kenya       |\\
| Stephen Cherono | Kenya       |\\ \\

3. \textbf{Step-by-Step Reasoning}:\\
- The statement claims that Kenya won 3 medals in the 3000 meters event at the 2001 Goodwill Games.\\
- Based on the execution result of SQL query, only two of the players that win medals are from Kenya.\\
- Therefore, the table context does not support the statement. The statement is false.\\
\\ \\
The final answer is \textbf{false}.\\\\
\#\# \textbf{Claim}\\
\textit{Claim}\\ \\

\#\# \textbf{Table Context}\\
\textit{Table}\\ \\

\#\# \textbf{Task}\\
You will answer the question based on the given context.You should reach a short-form answer after reasoning.
You are asked to answer the question in three steps.\\
1. Analyze the question and the given context. Make up a plan to answer the question.\\
2. Write one or more SQL to query the table for necessary information and output expected execution result.\\
3. Reason step-by-step to reach the final answer.\\ \\

\#\# \textbf{Answer}\\
\bottomrule
\end{tabular}
\caption{Prompt for generating responses for queries from Tabfact. GPT-3.5-turbo and GPT-4 generates responses following this example.}
\label{tab:tabfact_distill}
\end{table*}

\begin{table*}
\centering
\small
\setlength{\tabcolsep}{5pt}
\begin{tabular}{p{2.0\columnwidth}r}
\toprule
Check if the prediction answers the question correctly. For numerical answers, you should check if the predicted answer is approximately correct. For questions with multiple answers, you should check if all the predicted answers are correct. If the predicted answer is correct, return "Yes". Otherwise, return "No". The question, predicted answer, and gold answer are provided below.
\\ \\ 
\textbf{\#\# Question}\\
\textit{question}\\ \\
\textbf{\#\# Gold Answer}\\ 
\textit{gold answer}\\ \\
\textbf{\#\# Predicted Answer}\\
\textit{predicted answer}\\ \\
Does the prediction answer the question correctly? Yes/No\\
\textbf{\#\# Answer}
\\
\bottomrule
\end{tabular}
\caption{Prompt for question answering evaluation}
\label{tab:eval_prompt}
\end{table*}

\begin{table*}
\centering
\small
\setlength{\tabcolsep}{5pt}
\begin{tabular}{p{2.0\columnwidth}r}
\toprule
\textbf{Question Answering}\\ \\
\textbf{\#\# Question}\\
\textit{question}\\
\\ 
\textbf{\#\# Table}\\
\textit{table info}\\
\textit{table content}
\\ \\
\textbf{\#\# Sentence Context}\\
\textit{sentences}
\\ \\ 
\textbf{\#\# Task}\\
You will answer the question based on the given context.You should reach a short-form answer after reasoning. You are asked to answer the question in three steps.\\
1. Analyze the question and the given context. Make up a plan to answer the question.\\
2. Write one or more SQL to query the table for necessary information and output expected execution result.\\
3. Reason step-by-step to reach the final answer\\ \\
\textbf{\#\# Answer}\\
\midrule
\textbf{Fact Verification}\\ \\
\textbf{\#\# Claim}\\
\textit{claim}\\
\\ 
\textbf{\#\# Table}\\
\textit{table info}\\
\textit{table content}
\\ \\
\textbf{\#\# Sentence Context}\\
\textit{sentences}
\\ \\ 
\textbf{\#\# Task}\\
You will predict the claim’s label based on the given claim and context. There are 3 possible labels: SUPPORTS, REFUTES, and NOT ENOUGH INFO. SUPPORTS means the claim is supported by the context. REFUTES means the claim is refuted by the context. NOT ENOUGH INFO means the context does not provide enough information to the support or refute
the claim. You are asked to check the claim in three steps\\
1. Analyze the question and the given context. Make up a plan to answer the question.\\
2. Write one or more SQL to query the table for necessary information and output expected execution result.\\
3. Reason step-by-step to reach the final answer\\ \\
\textbf{\#\# Answer}
\\
\bottomrule
\end{tabular}
\caption{Prompt of short-form answer tasks for \textsc{ProTrix}. Table info includes page title, section title and caption. If no sentence context is provided, we discard the sentence context part in the prompt.}
\label{tab:eval_prompt_short}
\end{table*}

\begin{table*}
\centering
\small
\setlength{\tabcolsep}{5pt}
\begin{tabular}{p{2.0\columnwidth}r}
\toprule

\textbf{TableLlama} \& \textbf{GPT-3.5-turbo}

Below is an instruction that describes a task, paired with an input that provides further context. Write a response that appropriately completes the request.

\\
\#\#\# Instruction\\
This is a free-form table question answering task. The goal for this task is to answer the given question based on the given table.

\\
\#\#\# Input:\\

[TLE] The Wikipedia page title of this table is \textit{page title}. The Wikipedia section title of this table is \textit{section title}.\\

[TAB] \textit{table}
\\\\

\#\#\# Question:\\
\textit{question}

\\
\#\#\# Response:
\\
\midrule 
\textbf{Our model}\\
\#\# Question\\
\textit{question}

\\
\#\# Table Context
\\Page title: \textit{page title}
\\Section title:\textit{section title}
\\\textit{table}

\\\#\# Task
\\You will answer the question based on the given context. You should reach a free-form answer after reasoning. You are asked to answer the question in three steps.\\1. Analyze the question and the given context. Make up a plan to answer the question. \\2. Write one or more SQL to query the table for necessary information and output expected execution result.\\3. Reason step-by-step to reach the final answer.

\\
\#\#Answer:\\
\bottomrule
\end{tabular}
\caption{Prompt for evaluation on FetaQA}
\label{tab:fetaqa_prompt}
\end{table*}

\begin{table*}
\centering
\small
\setlength{\tabcolsep}{5pt}
\begin{tabular}{p{2.0\columnwidth}r}
\toprule

\#\# \textbf{Question}\\
How well did Peralta play at the 2015 CONCACAF Gold Cup?
\\ \\
\#\# \textbf{Table Context}\\
Page Title: Oribe Peralta\\
Caption: International goals \\
| Goal | Date | Venue | Opponent | Score | Result | Competition\\
...17 rows with goals in 2014 FIFA World Cup qualification\\
|18. | 9 July 2015 | Soldier Field, Chicago, United States | Cuba | 1-0 | 6-0 | 2015 CONCACAF Gold Cup \\
|19. | 9 July 2015 | Soldier Field, Chicago, United States | Cuba | 3-0 | 6-0 | 2015 CONCACAF Gold Cup \\
|20. | 9 July 2015 | Soldier Field, Chicago, United States | Cuba | 5-0 | 6-0 | 2015 CONCACAF Gold Cup \\
|21. | 26 July 2015 | Lincoln Financial Field, Philadelphia, United States | Jamaica | 3-0 | 3-1 | 2015 CONCACAF Gold Cup \\
...5 rows with goals in 2015 CONCACAF Cup, 2017 FIFA Confederations Cup\\
\midrule
\textbf{Gold answer}\\
Peralta played at the 2015 CONCACAF Gold Cup and scored a 6-0 defeat of Cuba in Chicago, as well as a 3-1 win against Jamaica in Philadelphia.
\\
\midrule
\textbf{GPT-3.5-turbo}\\

Peralta played very well at the 2015 CONCACAF Gold Cup, scoring a total of 8 goals in the tournament. \ding{55}\\
\midrule
\textbf{TableLlama}\\
Peralta scored twice in a 6-0 win over Cuba. \ding{55}\\
\midrule
\textbf{\textsc{ProTrix}}\\
1. Plan for answering the question\\
\quad - The question asks for an assessment of Peralta's performance at the 2015 CONCACAF Gold Cup. \\
\quad - To answer this, we need to filter the table for entries that are related to the 2015 CONCACAF Gold Cup and then analyze the results to determine if Peralta played well or not.\\ \\2. Write SQL and execute SQL\\ SQL\\SELECT Goal, Date, Venue, Opponent, Score, Result, Competition FROM International\_Goals WHERE Competition = '2015 CONCACAF Gold Cup'\\ \\Executed result:\\Goal | Date       | Venue                  | Opponent | Score | Result | Competition\\18.  | 9 July 2015 | Soldier Field, Chicago, United States | Cuba   | 1-0   | 6-0   | 2015 CONCACAF Gold Cup\\19.  | 9 July 2015 | Soldier Field, Chicago, United States | Cuba   | 3-0   | 6-0   | 2015 CONCACAF Gold Cup\\20.  | 9 July 2015 | Soldier Field, Chicago, United States | Cuba   | 5-0   | 6-0   | 2015 CONCACAF Gold Cup\\21. | 26 July 2015 | Lincoln Financial Field, Philadelphia, United States | Jamaica | 3-0 | 3-1 | 2015 CONCACAF Gold Cup\\\\
3. Step-by-Step Reasoning:\\
\quad $\bullet$ The SQL query filters the table to show only the goals scored by Peralta during the 2015 CONCACAF Gold Cup.\\
\sethlcolor{yellow}
\quad $\bullet$ From the execution result, we can see that \hl{Peralta scored three goals on 9 July 2015 against Cuba at Soldier Field, Chicago, United States and one goal on 26 July 2015 against Jamaica on 26 July 2015 at Lincoln Financial Field, Philadelphia, United States} \ding{52} \\
\quad  $\bullet$The results were 6-0 and 3-1, respectively, which indicates a dominant performance by Peralta and his team.\\
\quad$\bullet$ Therefore, Peralta played very well at the 2015 CONCACAF Gold Cup.\\ \\The final answer is that Peralta played very well at the 2015 CONCACAF Gold Cup.\quad \ding{52}\\
\bottomrule
\end{tabular}
\caption{An example of the responses of models to FetaQA}
\label{tab:fetaqa}
\end{table*}

\begin{table*}
\centering
\small
\setlength{\tabcolsep}{5pt}
\begin{tabular}{p{2.0\columnwidth}r}
\toprule
\#\# Question\\
who was the top chef?\\ \\ 

\#\# Table\\
Name | Age | Hometown | Occupation | Culinary P.O.V. | Eliminated\\
Damaris Phillips | 31 | Louisville, KY | Culinary Teacher | Modern Southern Food | Winner\\
Rodney Henry | 47 | Baltimore, MD | Pie Shop Owner | Pie Style | Second Place\\
Russell Jackson | 49 | San Francisco, CA | Underground Chef | Seven Culinary Sins | Third Place\\
Stacey Poon-Kinney | 34 | San Diego, CA | Restaurant Owner | Vintage with a Modern Twist | Week 10\\
Nikki Dinki | 29 | New York, NY | Food Blogger/Online Host | Semi-Vegetarian / Meat on the Side | Week 9\\
Chad Rosenthal | 37 | Ambler, PA | Restaurant Owner | Jewish BBQ Guy | Week 7\\
Chris Hodgson | 26 | Cleveland, OH | Chef/Restaurateur | Compassion for Food | Week 6\\
Viet Pham | 33 | Salt Lake City, UT | Chef/Restaurant Owner | The American Dream | Week 5\\
Connie Lovely Jackson" | 27 | Los Angeles, CA | Caterer | Party on a Plate | Week 4 / Week 8(Winner of Star Salvation)\\
Danushka Lysek | 37 | New York, NY | Private Chef/Model | Your Private Chef | Week 3\\
Andres Guillama | 26 | Waynesville, NC | Childhood Obesity Prevention Coach | Teaching Men to Cook | Week 2\\
Daniela Perez-Reyes | 28 | Haleiwa, HI | Bartender/Caterer | Peruvian Princess | Week 1\\
\midrule

\#\# Answer:\\
1. Plan for answering the question\\
- The question asks for the top chef, which implies we are looking for the winner of the competition.

- To find the top chef, we need to look for the name associated with the position of "Winner" in the table.
\\ \\
2. Write SQL and execute SQL \\
```sql\\
SELECT `Name` FROM w WHERE `Eliminated` = 'Winner'\\
```\\Expected Result:\\
```\\
Name\\
Damaris Phillips\\
```\\

3.Step-by-step Answer prediction\\
- The question is asking for the name of the top chef from the given table.

- Based on the execution result of the SQL query, the name associated with the position of "Winner" is Damaris Phillips.

- Therefore, the top chef according to the table is Damaris Phillips.

The final answer is Damaris Phillips.
\\
\bottomrule
\end{tabular}
\caption{An example from WikiTQ}
\label{tab:wikitab_case}
\end{table*}

\begin{table*}
\centering
\small
\setlength{\tabcolsep}{5pt}
\begin{tabular}{p{2.0\columnwidth}r}
\toprule
\#\# Claim\\
The houston texans score more point on november 16 , 2003 than on september 28 , 2003
\\ \\
\#\# Table\\
Caption: 2003 houston texans season\\
week | date | opponent | result | game site | record | tv time | attendance\\
1 | september 7 , 2003 | miami dolphins | w 21 - 20 | dolphin stadium | 1 - 0 | cbs 12:00 pm | 73010\\
2 | september 14 , 2003 | new orleans saints | l 10 - 31 | louisiana superdome | 1 - 1 | cbs 12:00 pm | 68390\\
3 | september 21 , 2003 | kansas city chiefs | l 14 - 42 | reliant stadium | 1 - 2 | cbs 12:00 pm | 70487\\
4 | september 28 , 2003 | jacksonville jaguars | w 24 - 20 | reliant stadium | 2 - 2 | cbs 12:00 pm | 70041\\
5 | - | - | - | - | - | - | \\
6 | october 12 , 2003 | tennessee titans | l 17 - 38 | lp field | 2 - 3 | cbs 12:00 pm | 68809\\
7 | october 19 , 2003 | new york jets | l 14 - 19 | reliant stadium | 2 - 4 | cbs 12:00 pm | 70623\\
8 | october 26 , 2003 | indianapolis colts | l 21 - 30 | rca dome | 2 - 5 | cbs 12:00 pm | 56132\\
9 | november 2 , 2003 | carolina panthers | w 14 - 10 | reliant stadium | 3 - 5 | fox 12:00 pm | 70052\\
10 | november 9 , 2003 | cincinnati bengals | l 27 - 34 | paul brown stadium | 3 - 6 | cbs 12:00 pm | 50437\\
11 | november 16 , 2003 | buffalo bills | w 12 - 10 | ralph wilson stadium | 4 - 6 | cbs 12:00 pm | 72677\\
12 | november 23 , 2003 | new england patriots | l 20 - 23 (ot) | reliant stadium | 4 - 7 | cbs 12:00 pm | 70719\\
13 | november 30 , 2003 | atlanta falcons | w 17 - 13 | reliant stadium | 5 - 7 | fox 12:00 pm | 70388\\
14 | december 7 , 2003 | jacksonville jaguars | l 0 - 27 | alltel stadium | 5 - 8 | cbs 12:00 pm | 43363\\
15 | december 14 , 2003 | tampa bay buccaneers | l 3 - 16 | raymond james stadium | 5 - 9 | cbs 12:00 pm | 65124\\
16 | december 21 , 2003 | tennessee titans | l 24 - 27 | reliant stadium | 5 - 10 | cbs 12:00 pm | 70758\\
17 | december 28 , 2003 | indianapolis colts | l 17 - 20 | reliant stadium | 5 - 11 | cbs 12:00 pm | 70680\\
\midrule
\#\# Answer:\\
Let's check the claim in 3 steps.\\ \\

1. Plan for verify the claim\\
- To verify the claim, we need to compare the points scored by the Houston Texans in the games on November 16, 2003, and September 28, 2003. We will write a SQL query to retrieve the result for these specific dates.
\\ \\
2. Write SQL and execute SQL\\
```sql\\
SELECT `date`, `result` FROM w WHERE `date` = 'november 16 , 2003' OR `date` = 'september 28 , 2003'\\
```Expected Result:\\

```\\
date            | result\\
november 16 , 2003 | w 12 - 10\\
september 28 , 2003 | w 24 - 20\\
```\\ \\

3. Step-by-step answer prediction\\
- The claim states that the Houston Texans scored more points on November 16, 2003, than on September 28, 2003.\\
- According to the SQL query result, the Houston Texans won with a score of 12-10 on November 16, 2003, and won with a score of 24-20 on September 28, 2003.\\
- The score on November 16, 2003, is less than the score on September 28, 2003.\\
- Therefore, the claim is refuted by the table.\\ \\

The answer is REFUTES.
\\
\bottomrule
\end{tabular}
\caption{An example from TabFact}
\label{tab:tabfact_case}
\end{table*}

\begin{table*}
\centering
\small
\setlength{\tabcolsep}{5pt}
\begin{tabular}{p{2.0\columnwidth}r}
\toprule
\#\# Claim\\
Here is a claim: In the natural state space with 75 actions, training does not converge within a reasonable training time. Does the following context support or refute the claim?
\\ \\
\#\# Table\\
Paper title: Guided Dialog Policy Learning: Reward Estimation for Multi-Domain Task-Oriented Dialog\\
Table caption: Table 5: Performance of different agents on the neural user simulator.\\
Method | VHUS Turns | VHUS Inform | VHUS Match | VHUS Success\\
ACER | 22.35 | 55.13 | 33.08 | 18.6\\
PPO | 19.23 | 56.31 | 33.08 | 18.3\\
ALDM | 26.90 | 54.37 | 24.15 | 16.4\\
GDPL | 22.43 | 52.58 | 36.21 | 19.7\\
\midrule

\#\# Answer:

1. Plan for answering the question\\
- The claim states that training does not converge within a reasonable training time in the natural state space with 75 actions.\\
- To answer the claim, we need to understand the training time and the number of actions in the state space.\\
- The table provides information about the performance of different agents, but it does not provide information about the training time or the number of actions.\\
- Since the table does not contain the necessary information to answer the claim, we cannot write an SQL query to extract this information from the table.\\
\\
2. Write SQL and execute SQL\\
- No SQL query can be written as the table does not contain the required information about the training time or the number of actions.\\
\\
3. Step-by-Step Reasoning:\\
- The claim is about the training time and the number of actions in the state space.\\
- The table provides information about the performance of different agents, but it does not provide information about the training time or the number of actions.\\
- Since the table does not contain the necessary information to answer the claim, we cannot confirm or refute the claim based on the provided context.\\
\\
Therefore, the answer is "NOT ENOUGH INFO".
\\
\bottomrule
\end{tabular}
\caption{An example from \textsc{SciTAB}}
\label{tab:scitab_case}
\end{table*}

\begin{table*}
\centering
\small
\setlength{\tabcolsep}{5pt}
\begin{tabular}{p{2.0\columnwidth}r}
\toprule
\#\# Claim\\
Delta Green was published by Arc Dream Publishing  which is a small role-playing game publishing company founded in 2002.
\\ \\
\#\# Table\\
Page Title: Delta Green\\
Caption: Delta Green\\
Designer(s) | Dennis Detwiller, Adam Scott Glancy, John Scott Tynes\\
Publisher(s) | Pagan Publishing Arc Dream Publishing Pelgrane Press (The Fall of DELTA GREEN)\\
Publication date | 1997 (Sourcebook) 2016 (Arc Dream Publishing) (Standalone RPG)2018 (The Fall of DELTA GREEN)\\
Genre(s) | Horror, Conspiracy fiction\\
System(s) | Call of Cthulhu, GUMSHOE\\
\\
\#\# Sentences:\\
Arc Dream Publishing: Arc Dream Publishing is a small role-playing game publishing company founded in 2002 by Pagan Publishing veteran Dennis Detwiller and editor Shane Ivey after the release of their first roleplaying game Godlike: Superhero Roleplaying in a World on Fire, 1936-1946 was published by Hobgoblynn Press.\\
Delta Green: The book was published in cooperation with Arc Dream Publishing.\\
Arc Dream Publishing: The name ARC DREAM comes from one of Dennis Detwiller's other roleplaying games—Delta Green—where it is a secret government project.\\
Delta Green: Delta Green is a setting for the Call of Cthulhu role-playing game created by Adam Scott Glancy, Dennis Detwiller, and John Scott Tynes, a.k.a. the Delta Green Partnership, of the Seattle gaming house Pagan Publishing.\\
Delta Green: In August 2011, Arc Dream Publishing and the Delta Green Partnership announced development of a standalone Delta Green role-playing game.\\
\midrule\\

\#\# Answer:\\
Let's check the claim in 3 steps.
\\ \\
1. Plan for verify the claim\\
- To verify the claim, we need to confirm that Delta Green was published by Arc Dream Publishing and that Arc Dream Publishing is a small role-playing game publishing company founded in 2002.\\
- We can use the table to verify the publisher of Delta Green and the sentence context to confirm the founding year of Arc Dream Publishing.\\ \\

2. Write SQL and execute SQL\\
```sql\\
SELECT Publisher FROM w\\
```\\
Expected result:\\
```\\
Publisher\\
Pagan Publishing\\
Arc Dream Publishing\\
Pelgrane Press (The Fall of DELTA GREEN)\\
```\\
\\
3. Step-by-step answer prediction\\
- The claim states that Delta Green was published by Arc Dream Publishing, which is a small role-playing game publishing company founded in 2002.\\
- The SQL query result shows that Delta Green was published by Pagan Publishing, Arc Dream Publishing, and Pelgrane Press. This supports the claim that Arc Dream Publishing is one of the publishers of Delta Green.\\
- The sentence context Arc Dream Publishing confirms that Arc Dream Publishing is a role-playing game publishing company founded in 2002.\\
- The sentence context Delta Green supports the claim that Delta Green is a setting for the Call of Cthulhu role-playing game.\\
\\
Therefore, the answer is SUPPORTS.\\
\bottomrule
\end{tabular}
\caption{An example from FEVEROUS}
\label{tab:feverous_case}
\end{table*}

\begin{table*}
\centering
\small
\setlength{\tabcolsep}{5pt}
\begin{tabular}{p{2.0\columnwidth}r}
\toprule
\#\# Question

when did the judge born August 4 , 1791 in Sharon , Litchfield County , Connecticut leave office ?

\\
\#\# Table\\
Name | Took office | Left office | Party\\
Freeborn G. Jewett | July 5 , 1847 | December 31 , 1849 | Democratic\\
Greene C. Bronson | January 1 , 1850 | April 1851 | Democratic/ Anti-Rent\\
Charles H. Ruggles | April 1851 | December 31 , 1853 | Democratic\\
Addison Gardiner | January 1 , 1854 | December 31 , 1855 | Democratic/Anti-Rent\\
Hiram Denio | January 1 , 1856 | December 31 , 1857 | Democratic\\
Alexander S. Johnson | January 1 , 1858 | December 31 , 1859 | Democratic\\
George F. Comstock | January 1 , 1860 | December 31 , 1861 | American\\
Samuel L. Selden | January 1 , 1862 | July 1 , 1862 | Democratic\\
Hiram Denio | July 1 , 1862 | December 31 , 1865 | Democratic\\
Henry E. Davies | January 1 , 1866 | December 31 , 1867 | Republican / American\\
William B. Wright | January 1 , 1868 | January 12 , 1868 | Union\\
Ward Hunt | January 12 , 1868 | December 31 , 1869 | Republican\\
Robert Earl | January 1 , 1870 | July 4 , 1870 | Democratic\\

\\
\#\# Sentences:\\
Freeborn G. Jewett: Freeborn Garrettson Jewett ( August 4 , 1791 in Sharon , Litchfield County , Connecticut - January 27 , 1858 in Skaneateles , Onondaga County , New York ) was an American lawyer and politician who served as a U.S. Representative from New York and was the first Chief Judge of the New York Court of Appeals .\\
Greene C. Bronson: Greene Carrier Bronson ( November 17 , 1789 in Simsbury , Hartford County , Connecticut - September 3 , 1863 in Saratoga , New York ) was an American lawyer and politician from New York .\\
Samuel L. Selden: Samuel Lee Selden ( October 12 , 1800 Lyme , New London County , Connecticut - September 20 , 1876 Rochester , Monroe County , New York ) was an American lawyer and politician from New York .
\\
\midrule

\#\# Answer:
1. Plan for answering the question\\
- The question asks for the date when a judge named Freeborn G. Jewett left office.\\
- To answer the question, we need to find the row in the table that corresponds to Freeborn G. Jewett and then select the date he left office.

\\
2. Write SQL and execute SQL \\
```sql\\
SELECT `Left office` FROM w WHERE `Name` = 'Freeborn G. Jewett'\\
```Expected Result:\\
```\\
Left office\\
1849-12-31\\
```\\

\\
3.Step-by-step Answer prediction:\\
- The question is asking for the date when a specific judge, Freeborn G. Jewett, left office.\\
- Based on the execution result of the SQL query, Freeborn G. Jewett left office on December 31, 1849.\\
- Therefore, the date the judge left office is December 31, 1849.\\

The final answer is December 31, 1849.\\
\bottomrule
\end{tabular}
\caption{An example from HybridQA}
\label{tab:hybridqa_case}
\end{table*}

\begin{table*}
\centering
\small
\setlength{\tabcolsep}{5pt}
\begin{tabular}{p{2.0\columnwidth}r}
\toprule
\#\# Question\\
What is the average proportion of cost of revenue as a percentage of the total revenue in 2018 and 2019?\\

\#\# Table\\
Fiscal Years |  |  | \\
 | 2019 | 2018 | 2017\\
Statements of Operations: |  |  | \\
Revenue | 100\% | 100\% | 100\%\\
Cost of revenue | 43\% | 50\% | 55\%\\
Gross profit | 57\% | 50\% | 45\%\\
Operating expenses: |  |  | \\
Research and development | 120\% | 79\% | 79\%\\
Selling, general and administrative | 86\% | 79\% | 81\%\\
Loss from operations | (149)\% | (108)\% | (115)\%\\
Interest expense | (3)\% | (1)\% | (1)\%\\
Interest income and other expense, net | 2\% | 1\% | —\%\\
Loss before income taxes | (150)\% | (108)\% | (116)\%\\
Provision for income taxes | 1\% | 1\% | 1\%\\
Net loss | (151)\% | (109)\% | (117)\%\\\\

\#\# Sentences\\
The following table sets forth the percentage of revenue for certain items in our statements of operations for the periods indicated:\\
Impact of inflation and product price changes on our revenue and on income was immaterial in 2019, 2018 and 2017.\\
\midrule

\#\# Answer\\ \\
1. Plan for answering the question\\
- The question asks for the average proportion of cost of revenue as a percentage of the total revenue for the years 2018 and 2019.

- To find the average, we need to calculate the proportion for each year and then find the mean of these two values.

- We will write an SQL query to select the cost of revenue as a percentage of revenue for the years 2018 and 2019.
\\ \\ 
2. Write SQL and execute SQL \\
```sql\\
SELECT `Fiscal Years`, `Cost of revenue` FROM Table WHERE `Fiscal Years` IN ('2019', '2018')\\
```\\
Expected result:\\
```\\
| Fiscal Years | Cost of revenue |\\
| 2019        | 43\%             |\\
| 2018        | 50\%             |\\
```\\
\\
3. Step-by-Step Reasoning:\\
- The SQL query gives us the cost of revenue as a percentage of revenue for the years 2019 and 2018.

- To find the average, we calculate the mean of the two values: \red{(0.43 + 0.50) / 2 = 0.48}.

- Therefore, the average proportion of cost of revenue as a percentage of the total revenue for 2018 and 2019 is 48\%.
\\ \\
The final answer is 48\%.\\
\midrule
\#\#Gold Answer\\
46.5\%\\
\bottomrule
\end{tabular}
\caption{An error case from responses to TATQA. The error source is highlighted in red.}
\label{tab:error_tat}
\end{table*}

\begin{table*}
\centering
\small
\setlength{\tabcolsep}{5pt}
\begin{tabular}{p{2.0\columnwidth}r}
\toprule
\#\# Claim\\
Campina Grande had three HDIs  between 1991 and 2000, one of which is Value.
\\ \\
\#\# Table\\
Page Title: Campina Grande\\
Caption: UNHD (2000)\\
HDI | 1991 | 2000\\
Value | 0.615 | 0.678\\
Life expectancy | 0.585 | 0.641\\
Education | 0.741 | 0.844\\
Total | 0.647 | 0.721\\
\\
\#\# Sentences\\
Campina Grande: Campina Grande is the second most populous Brazilian city in the State of Paraiba after Joao Pessoa, the capital.  

Roman Catholic Diocese of Campina Grande: The Roman Catholic Diocese of Campina Grande (Latin: Dioecesis Campinae Grandis) is a suffragan Latin diocese in the Ecclesiastical province of Paraiba, in northeastern Brazil.

Campina Grande: It is considered one of the main industrial, technological and educational centers in the northeastern region of Brazil.

Campina Grande: For a long time Campina Grande has had a strong technological focus.

Campina Grande: At the time, it was the second biggest producer in the world.
\\
\midrule
\#\# Answer:\\
Let's check the claim in 3 steps.\\
1. Plan for verify the claim\\
- To verify the claim, we need to check the HDI values for Campina Grande for the years 1991 and 2000. The table provides the necessary data to verify this part of the claim.\\
- We do not need to use the sentences for this claim as the table provides the required information.
\\ \\
2. Write SQL and execute SQL\\
```sql\\
SELECT `HDI`, `1991`, `2000` FROM w\\
```Expected Result:\\
```\\
HDI | 1991 | 2000\\
Value | 0.615 | 0.678\\
Life expectancy | 0.585 | 0.641\\
Education | 0.741 | 0.844\\
Total | 0.647 | 0.721\\
```\\
\\
3.Step-by-step Answer prediction

- The claim states that Campina Grande had three HDIs between 1991 and 2000, and one of them is Value.\\
- The SQL query result shows that there are \red{four HDIs} listed for Campina Grande in the years 1991 and 2000: Value, Life expectancy, Education, and \red{Total}.\\
- The claim specifies "Value" as one of the HDIs, but the table shows that there are multiple values (pun intended), which refutes the claim that there was only one HDI.\\
\\
Therefore, the answer is REFUTES.\\
\midrule
\#\#Gold Answer\\
SUPPORTS\\
\bottomrule
\end{tabular}
\caption{An error case from responses to FEVEROUS. The error source is highlighted in red.}
\label{tab:error_feverous}
\end{table*}

\end{document}